\documentclass[letterpaper]{article} 
\usepackage{aaai2026}
\usepackage{times}  
\usepackage{helvet}  
\usepackage{courier}  
\usepackage[hyphens]{url}  
\usepackage{multirow} 
\usepackage{graphicx} 
\usepackage{booktabs}      
\usepackage{subcaption}
\usepackage{multirow}      
\usepackage{graphicx}      
\usepackage{caption}       
\usepackage{colortbl}
\usepackage[utf8]{inputenc}
\usepackage{tcolorbox}
\tcbuselibrary{listingsutf8}
\usepackage{tcolorbox}
\tcbuselibrary{listings, skins, breakable}
\usepackage[frozencache,cachedir=.]{minted}
\usepackage{natbib}  
\usepackage{caption} 
\usepackage{amsfonts}
\usepackage{amsmath}
\usepackage{amssymb}
\usepackage{amsthm}
\usepackage{algpseudocode}
\usepackage[linesnumbered,ruled,vlined]{algorithm2e} 
\usepackage{booktabs}
\usepackage{tcolorbox}
\usepackage{listings}
\usepackage{xcolor}

\usepackage{xcolor}
\usepackage{listings}
\usepackage{tcolorbox}
\usepackage{courier}
\usepackage{inconsolata}  

\usepackage{minitoc} %

\noptcrule

\usepackage{etoc}
\usepackage{enumitem} 

\usepackage{listings}
\DeclareCaptionStyle{ruled}{labelfont=normalfont,labelsep=colon,strut=off} 
\floatstyle{ruled}
\newfloat{listing}{tb}{lst}{}
\floatname{listing}{Listing}

\title{
Pareto-Grid-Guided Large Language Models for Fast and High-Quality Heuristics Design in Multi-Objective Combinatorial Optimization
}

\author {
    Minh Hieu Ha\textsuperscript{\rm 1}\thanks{Equal contribution.},
    Hung Phan\textsuperscript{\rm 1}\footnotemark[1],
    Tung Duy Doan\textsuperscript{\rm 1},
    Tung Dao\textsuperscript{\rm 1},
    Dao Tran\textsuperscript{\rm 2},
    Huynh Thi Thanh Binh\textsuperscript{\rm 1}
}

\affiliations {
\textsuperscript{\rm 1}Hanoi University of Science and Technology, Vietnam\\
\textsuperscript{\rm 2}FPT Software AI Center, Vietnam\\

\{hieu.hm220026, hung.pd214903, tung.dd224906, tung.dv242050M\}@sis.hust.edu.vn, \\
daotc2@fpt.com, binhht@soict.hust.edu.vn
}

\usepackage[acronym]{glossaries}
\usepackage{subfiles}
\newacronym{llm}{LLM}{Large Langue Model}
\newacronym{moo}{MOO}{Multi-objective Optimization}
\newacronym{cop}{COP}{Combinatorial Optimization Problem}
\newacronym{tsp}{TSP}{Traveling Salesman Problem}
\newacronym{vrp}{VRP}{Vehicle Routing Problem}

\newacronym{pfg-moea}{PFG-MOEA}{Pareto Front Grid guided Multi-objective Evolutionary Algorithm}

\newacronym{pfg}{PFG}{Pareto Front Grid}

\newacronym{mocop}{MOCOP}{Multi-objective combinatorial optimization problems}
\newacronym{llms}{LLMs}{Large Language Models}

\newacronym{semo}{SEMO}{Simple Evolutionary Multiobjective Optimization}

\newacronym{moeas}{MOEAs}{Multi-objective evolutionary algorithms}

\newacronym{mpage}{MPaGE}{\textbf{\underline{M}}ulti-heuristics for MOCOP via \textbf{\underline{Pa}}reto-\textbf{\underline{G}}rid-guided \textbf{\underline{E}}volution of LLMs}

\usepackage{xcolor}
\usepackage{subcaption}
\usepackage{soul}

\usepackage{enumitem}

\begin{document}
\maketitle
\begin{abstract}
\gls{mocop} frequently arise in practical applications that require the simultaneous optimization of conflicting objectives. Although traditional evolutionary algorithms can be effective, they typically depend on domain knowledge and repeated parameter tuning, limiting flexibility when applied to unseen \gls{mocop} instances. Recently, integration of \gls{llms} into evolutionary computation has opened new avenues for automatic heuristic generation, using their advanced language understanding and code synthesis capabilities. Nevertheless, most existing approaches predominantly focus on single-objective tasks, often neglecting key considerations such as runtime efficiency and heuristic diversity in multi-objective settings. 
To bridge this gap, we introduce \gls{mpage}, a novel enhancement of the \gls{semo} framework that leverages \gls{llms} and \gls{pfg} technique. 
By partitioning the objective space into grids and retaining top-performing candidates to guide heuristic generation, \gls{mpage} utilizes LLMs to prioritize heuristics with semantically distinct logical structures during variation, thus promoting diversity and mitigating redundancy within the population. 
Through extensive evaluations, \gls{mpage} demonstrates superior performance over existing LLM-based frameworks, and achieves competitive results to traditional \gls{moeas}, with significantly faster runtime. 
Our code is available at \url{https://github.com/langkhachhoha/MPaGE}.
\end{abstract}

%

\section{Introduction}
\acrfull{mocop} commonly arise in real-world applications such as vehicle routing, production planning, where multiple conflicting objectives must be optimized simultaneously over a large, discrete solution space \citep{turkyilmaz2020research, liu2020multi, hung2025}. Unlike single-objective optimization, \gls{mocop} aims to approximate the Pareto front, which captures trade-offs among non-dominated solutions.

Due to the NP-hard nature of these problems, exact algorithms are often impractical, resulting in the widespread use of heuristic and metaheuristic methods such as NSGA-II, MOEA/D, and MOEA/D-DE \citep{deb2002fast, zhang2007moea, li2008multiobjective}. Despite their success, these methods typically rely on domain-specific knowledge and extensive iterative search. Several neural approaches have been proposed for \gls{mocop}, aiming to automatically learn heuristics and improve adaptability \citep{ zhang2022meta, hieu2024, fan2024con}. However, these methods often require retraining for different problem sizes, demand substantial computational resources, and struggle to generalize to problems with unseen input formats.

Recently, \acrfull{llms} have shown strong capabilities in automatic heuristic design, offering a new paradigm for optimization algorithm development \citep{wu2024evolutionary, liu2025large, novikov2025alphaevolve}. By leveraging their language understanding and code generation abilities, \gls{llms} can produce heuristics and the corresponding implementations with minimal human intervention. Recent approaches integrate \gls{llms} with evolutionary strategies to iteratively evolve effective problem-solving programs \citep{liu2024eoh, ye2024reevo, romera2024mathematical, van2024llamea}. While achieving competitive performance and often surpassing traditional methods, most existing LLM-based evolution frameworks primarily target single-objective problems, with limited exploration of their applicability to multi-objective settings. 
The inherent complexity and trade-off structure of \gls{mocop} introduce significant challenges that require tailored and robust design strategies. 
  To address this, Huang et al. \citep{huang2025autonomous} propose an LLM-based framework for discovering and refining evolutionary operators suited to diverse \gls{mocop}. However, runtime efficiency, which is crucial for the real-world deployment of \gls{mocop} solvers, remains an underexplored aspect of LLM-driven heuristic design. MEoH~\citep{yao2025meoh} considers both optimality and efficiency within a multi-objective evolutionary framework. Additionally, prior LLM-based methods tend to produce populations of algorithms with similar operational logic and slight representation difference, limiting both population diversity and the creativity of \gls{llms}.

To address the aforementioned issues, we propose \gls{mpage}, a novel framework designed to solve \gls{mocop} while simultaneously discovering a pareto front of LLM-generated heuristics that jointly optimize solution quality and runtime efficiency, with an explicit emphasis on promoting heuristic diversity. Our approach curates heuristic algorithms for the \acrfull{semo} paradigm, leveraging \acrfull{pfg} to guide the design of LLM-based variation heuristics by partitioning the objective space into grids, retaining leading individuals from promising regions, and enhancing both solution quality and search efficiency.
From these regions, \gls{mpage} builds a pool of elitist candidates and employs \gls{llms} to assess their semantic structures, clustering them into groups of similar logic. Variation is then performed with respect to these clusters, promoting semantic diversity and mitigating redundancy within the heuristic population. 
To the best of our knowledge, this is the first comprehensive evaluation of LLM-generated heuristics on standard \gls{mocop}, addressing both solution quality, computational efficiency and semantic diversity.
Our main contributions are as follows:
\begin{itemize}
\item We introduce \gls{mpage}, the first framework to systematically combine \gls{llms} with the \gls{semo} paradigm and \gls{pfg}, aiming to solve \gls{mocop} by jointly optimizing runtime, solution quality, and maintaining semantic diversity.
\item We leverage LLMs to verify the logical structure of heuristics and perform cross-cluster recombination, thereby enhancing diversity and reducing redundancy through logically dissimilar variations.
\item We conduct extensive experiments on standard \gls{mocop} benchmarks, demonstrating consistent improvements in runtime efficiency, solution quality, and semantic diversity over LLM-based baselines and traditional \gls{moeas}.

\end{itemize}

\vspace{-1em}
\section{Related Works}
\vspace{-0.2em}
\subsection{Multi-objective Optimization Algorithms}
Solving \gls{mocop} often involves extending metaheuristics to handle multiple objectives in discrete domains.
NSGA-II \citep{deb2002fast} and MOEA/D \citep{zhang2007moea} remain popular for effectively maintaining Pareto front diversity. Local search techniques like Pareto Local Search (PLS) \citep{paquete2004pareto}, \gls{semo} \citep{laumanns2004runtime} improve exploitation by iteratively exploring non-dominated neighbors.
More recent neural approaches aim to learn the entire Pareto set, as in PMOCO \citep{lin2022pareto}, or boost diversity through dual mechanisms, as in NHDE \citep{chen2023nhde}. Despite their success in solving \gls{mocop}, these methods still rely on handcrafted components, struggle to generalize across diverse problem settings, and incur substantial computational costs.
\vspace{-1.5em}
\subsection{LLMs for Heuristic Design}
\gls{llms} are increasingly used to automate heuristic design in combinatorial optimization. Core approaches like EoH, AEL, and FunSearch employ evolutionary frameworks to generate, combine, and refine heuristics in natural language or code, outperforming traditional methods on TSP, bin packing, and scheduling tasks \citep{liu2024eoh, liu2023ael, romera2024funsearch}. Similarly, HSEvo incorporates harmony search principles into the LLM-driven generation process to promote diversity and adaptability in solutions \citep{dat2025hsevo}.
Other work combines LLMs with synthesis techniques like MCTS for guided search \citep{zheng2025mcts}. While promising, most of these methods primarily focus on single-objective tasks, with limited attention to \gls{mocop}, neglects other practical criteria like efficiency and complexity \citep{gutjahr2012runtime}.
\subsection{Multi-Objective Optimization with LLMs}
Recent LLM-based heuristics have been applied to meta-heuristic design. Huang et al. \citep{huang2025autonomous} used LLMs to generate crossover and mutation operators for multi-objective optimization, later extending this to evolutionary multitasking with LLM-designed knowledge transfer models \citep{huang2024advancing}. Nevertheless, these methods overlook the computational cost of generating and evaluating heuristics, a key challenge in \gls{moeas}. 
MEoH, REMoH \citep{yao2025meoh, fornies2025remoh} present a multi-objective evolutionary framework that optimizes multiple performance metrics, such as optimality and efficiency, to discover trade-off algorithms in a single run. However, they struggle to distinguish heuristics with similar logic but different implementations, reducing diversity on the pareto front and hindering multi-objective exploration.
\vspace{-1em}
\section{Preliminary}
\vspace{-0.3em}
\subsection{Multi-Objective Combinatorial Optimization}
Generally, we consider \gls{mocop} defined as:
\(
  \min_{x \in \mathcal{X}}\; f(x) = (f_1(x), \ldots, f_M(x)),
\)
where $\mathcal{X}$ is a finite or discrete feasible set, and $f \colon \mathcal{X} \to \mathbb{R}^M$ maps each solution to $M$ objectives to be minimized. Trade-offs among objectives are characterized by Pareto optimality, as defined below, and solution quality is commonly assessed using the hypervolume (HV) indicator \citep{zitzler1999multiobjective}.

\noindent\textbf{Definition 1 (Pareto dominance).} For solutions \(x^a, x^b \in \mathcal{X}\), \(x^a\) dominates \(x^b\), denoted \(x^a \prec x^b\), if \(f_i(x^a) \leq f_i(x^b)\) for all \(i \in \{1, \ldots, M\}\), and there exists some \(j\) such that \(f_j(x^a) < f_j(x^b)\).
\\
\noindent\textbf{Definition 2 (Pareto optimality).} A solution $x^* \in \mathcal{X}$ is Pareto optimal if no $x \in \mathcal{X}$ satisfies $x \prec x^*$. The set of all such solutions forms the Pareto set ($\mathrm{PS}$), and Pareto front $\mathrm{PF} = \{{f}(x) \mid x \in \mathrm{PS}\}$.

\begin{figure*}[!ht]          
  \centering
  \includegraphics[width=\textwidth]{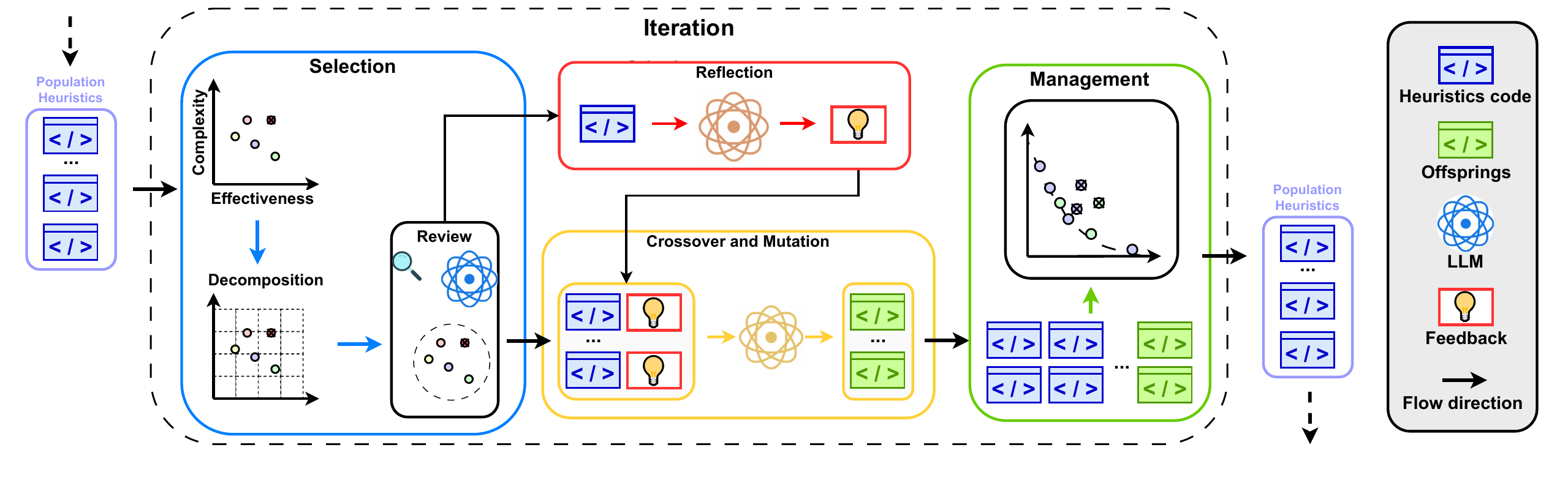}
  \vspace{-1em}
  \caption{\small Overview of the proposed method. The heuristics are first represented in a feature space, and then partitioned into grid cells based on their objective values. Potential parent heuristics are then chosen with the assistance of LLM-based review from grids. Crossover and mutation operations are applied, guided by informative feedback from LLM-based reflection to enhance population diversity. Finally, non-dominated individuals are retained to form the next generation for the subsequent iteration.}
  \label{fig:pfgllm-overview}
  \vspace{-1.5em}
\end{figure*}

\vspace{-0.8em}
\subsection{Simple Evolutionary Multiobjective Optimization}
\acrfull{semo} is a minimalist yet effective \gls{moeas}, widely used in theoretical analyses of runtime performance \citep{laumanns2004runtime}.
Similar to Pareto Local Search (PLS) \citep{paquete2004pareto}, SEMO maintains an archive of nondominated solutions. As shown in Algorithm \ref{alg:semo}, each iteration selects a random solution from the archive, explores one of its neighbors, and updates the archive if a new nondominated solution is discovered. Unlike PLS, which evaluates all neighbors, SEMO samples only one neighbor per iteration, which may limit its anytime performance. See appendix B for detailed insights compared to other frameworks.

\vspace{-1em}




    
    


\let\oldnl\nl            
\renewcommand{\nl}{\relax}  

\vspace{-1pt}

\begin{algorithm}[ht]
\caption{\small{Simple Evolutionary Multiobjective Optimization}}
\label{alg:semo}
\small


$s \gets \textit{rand\_generation}()$ \tcp*{Initialize randomly a solution}

$A \gets \{s\}$\;

\Repeat{$stop\_condition()$}{
    $s \gets \textbf{\textit{selection}}(A)$ \tcp*{Random $s$ from $A$}
    
    $s' \gets \textbf{\textit{neighborhood\_exploration}}(s)$ \tcp*{Select a neighbor of $s$ randomly}
    
    \If(\tcp*[f]{Check dominance}){$\nexists a \in A : a \prec s'$}{
        $A \gets A \cup \{s'\} \setminus \{a \in A \mid s' \prec a\}$\;
    }
}

\Return $A$\;
\end{algorithm}

\let\nl\oldnl  

\section{Methodology}
\subsection{Problem Formulation}


We formalize heuristics design for \gls{mocop} as a Language Multi-Criteria Heuristic Design (LMHD) problem, aiming to discover heuristics that balance evaluation criteria reflecting various aspects of heuristic behavior, such as effectiveness, efficiency, or generalization across problems. The key components are introduced as follows:

\noindent\textbf{Multi-Criteria Evaluation Function. }  
We define $E: \mathcal{H} \rightarrow \mathbb{R}^M$ to evaluate each heuristic $h \in \mathcal{H}$ over $M$ criteria, capturing its expected performance and key behavioral aspects on a given \gls{mocop}.

\noindent\textbf{Language Multi-Criteria Heuristic Design. }  
Language Multi-Criteria Heuristic Design (LMHD) is a variant of hyper-heuristic that leverages \gls{llms} to generate a diverse set of heuristics. The goal is to discover heuristics that balance multiple criteria simultaneously.
Given the heuristic space $\mathcal{H}$, the evaluation function  
\(
E(h) = \left(e_1(h), e_2(h), \ldots, e_M(h)\right)
\)
assigns to each heuristic $h$ a vector of $M$ criteria to be minimized. The aim is to approximate the Pareto front of nondominated heuristics in $\mathcal{H}$.

Our goal is to design MOCOP-specific heuristics for the selection and neighborhood exploration steps, corresponding to lines 4 and 5 in Algorithm~\ref{alg:semo}.
The proposed LMHD takes problem specifications as input and generates heuristics optimized over two practical and complementary criteria: solution quality and computational efficiency. Specifically, we define:
\begin{equation}
E: \mathcal{H} \rightarrow \mathbb{R}^2, \quad E(h) = \left(e_1(h), e_2(h)\right),
\end{equation}
where $e_1$ is the average negative hypervolume across a set of instances, measuring solution quality, and $e_2$ is the total running time. These two criteria reflect critical trade-offs in real-world applications, where high performance must be balanced with efficiency.
Each heuristic $h \in \mathcal{H}$ is a complete algorithm that, given the current population $A$, returns a new candidate solution $s'$ by internally performing both selection and neighborhood generation, as in Algorithm 1:
\begin{equation}
h: 2^{\mathcal{|S|}} \rightarrow \mathcal{S}, \quad s' = h(A),
\end{equation}
where $\mathcal{S}$ denotes the space of feasible solutions.

\subsection{Overview}
\gls{mpage} integrates \gls{llms} into a multi-phase evolutionary framework to design diverse and effective heuristics, as illustrated in Figure~\ref{fig:pfgllm-overview}. 
The process begins by initializing a population of heuristics tailored for SEMO paradigm. Each heuristic is represented in a feature space defined by solution quality and running time, and the population is progressively refined until the stopping criterion is met, yielding a set of non-dominated heuristics.
At each iteration, PFG generates grids that partition the objective space, thereby grouping heuristics into distinct regions (Section~\ref{Pareto Front Grid Generation}). A pool of elitism heuristics is selected from these grids, and an LLM-based review analyzes the logical semantics of candidate heuristics to cluster them into behaviorally similar groups (Section~\ref{Logical Assessment Mechanism}). Subsequently, search operators such as crossover and mutation are applied to these clusters, guided by informative feedback from LLM-based reflection, to generate new offspring. These offspring are added to the population, and non-dominated individuals are selected to form the next generation for the subsequent iteration (Section~\ref{Automatic Heuristic generation}).
\vspace{-0.5em}

\subsection{Pareto Front Grid mechanism}
\label{Pareto Front Grid Generation}

Dominance-based approaches like NSGA-II \citep{deb2002fast} and decomposition-based methods such as MOEA/D \citep{zhang2007moead} often struggle to balance convergence and diversity in tracking the true Pareto Fronts, particularly in heuristic design scenarios where objectives like runtime and solution quality exhibit irregular or non-uniform distributions. To address these limitations, we adopt the Pareto Front Grid (PFG) approach \cite{xu2023} to achieve a better balance between convergence and diversity.
Guiding the search using leading solutions in \gls{pfg} helps focus on promising regions, improving solution quality and efficiency while reducing redundancy.

\noindent\textbf{PFG Generation:} Motivated by \cite{xu2023}, let the objective space be denoted by \( \mathcal{Z} \subset \mathbb{R}^2 \). At generation \( t \), let \( \mathcal{P}^{(t)} = \{{E}({h}_i) \in \mathbb{R}^2 \mid {h}_i \in \mathcal{H}^{(t)} \} \) be the set of objective vectors of the current population, where each \({h}_i \) is a heuristic algorithm and \( \mathcal{H}^{(t)}\) denotes the heuristic space. 
We define the \textit{ideal} and \textit{nadir} points for each objective $j \in \{1, 2\}$ as  
\( z^*_j = \min_{h \in \mathcal{P}^{(t)}} e_j(h) \) and  
\( z^n_j = \max_{h \in \mathcal{P}^{(t)}} e_j(h) \),  
where \( e_j(h) \) denotes the $j$-th objective value of heuristic $h$. The objective space is scaled via min-max normalization:
\begin{equation}
\tilde{{E}} = \left( \frac{e_1 - z^*_1}{z^n_1 - z^*_1}, \frac{e_2 - z^*_2}{z^n_2 - z^*_2} \right)^\top.
\end{equation}

To structure the population in the normalized objective space, we partition \( [0,1]^2 \) into a grid of cells of side length \( \delta_1 > 0 \), \( \delta_2  > 0\) along each objective axis, respectively, depending on the number of segments along each dimension.
For each solution \( {h}_i \), assign its objective vector \( \tilde{{E}}({h}_i) \) to a grid cell \( G({h}_i) \in \mathbb{N}^2 \), and define the \gls{pfg} mapping as:

\begin{equation}
G({h}_i) = \left( 
\left\lfloor \frac{e_1({h}_i)}{\delta_1} \right\rfloor,\ 
\left\lfloor \frac{e_2({h}_i)}{\delta_2} \right\rfloor 
\right) \in \mathbb{N}^2.
\end{equation}

This discretization forms a structured grid over the objective space, grouping solutions into distinct regions. We define the following mapping:
\begin{equation}
\mathcal{G} : \mathbb{N}^2 \to 2^{\mathcal{|H|}}, \quad 
\mathcal{G}(g) = \left\{{h}_i \in \mathcal{H}^{(t)} \mid G({h}_i) = g \right\}.
\end{equation}
where each cell \( g \) contains all solutions whose objective vectors fall within that grid region.
For each non-empty cell \( g \in \operatorname{dom}(\mathcal{G}) \), retain a representative subset \( \mathcal{R}_g \subset \mathcal{G}(g) \), selected using non-dominated sorting within the cell. The union of all representatives yields the elite set:
\begin{equation}
\label{eq:union}
\mathcal{E}^{(t)} = \bigcup_{g} \mathcal{R}_g.
\end{equation} 
\noindent\textbf{PFG Selection:} 
To facilitate reproduction, the population is organized into a grid, where heuristics exhibiting similar running time or solution quality are placed in neighboring cells. This spatial organization encourages crossover between heuristics with related characteristics, promoting the inheritance and refinement of useful traits.

Selection for mating is performed over a pool \( \mathcal{P} \)
formed using the grid structure of the objective space.
With probability \( \epsilon \), a group of elitism candidates is selected from a set of adjacent grid cells. Specifically, a grid cell \( g \) is randomly sampled, and the pool is formed as the union of \( \mathcal{G}(g) \) and its neighboring cells \( \mathcal{G}(g') \), where \( g' \) denotes the cells adjacent to \( g \) along both objective axes, as shown in Figure \ref{fig:selectionpfg}. Otherwise, the parents are selected from entire \( \mathcal{E}^{(t)} \). Formally:
\begin{equation}
\label{eq:selection}
\mathcal{P} =
\begin{cases}
\mathcal{G}(g) \cup \mathcal{G}(g'), & \text{if } \mathcal{U}[0,1] < \epsilon,\\
\mathcal{E}^{(t)}, & \text{otherwise},
\end{cases}
\end{equation}
This hybrid strategy balances local exploitation and global exploration, promoting diversity while guiding search toward the Pareto front.
See Appendix C for details.
\vspace{-0.5em}

\subsection{Semantic clustering}
\label{Logical Assessment Mechanism}
Previous methods struggle to maintain effective diversity in multi-objective heuristic design, especially for complex \gls{mocop}. The standard \gls{moeas} relies on Pareto dominance in the objective space \( f(x) \), which is unreliable due to the stochastic nature of the heuristic performance. MEoH addresses this by introducing a dominance-dissimilarity measure based on Abstract Syntax Trees (ASTs) \citep{yao2025meoh, neamtiu2005understanding}. However, ASTs fail to capture semantic similarity between heuristics with similar logic but different implementations, resulting in redundant individuals and reduced diversity.
\begin{figure}[!h]
    \centering
    \includegraphics[width=0.99\linewidth]{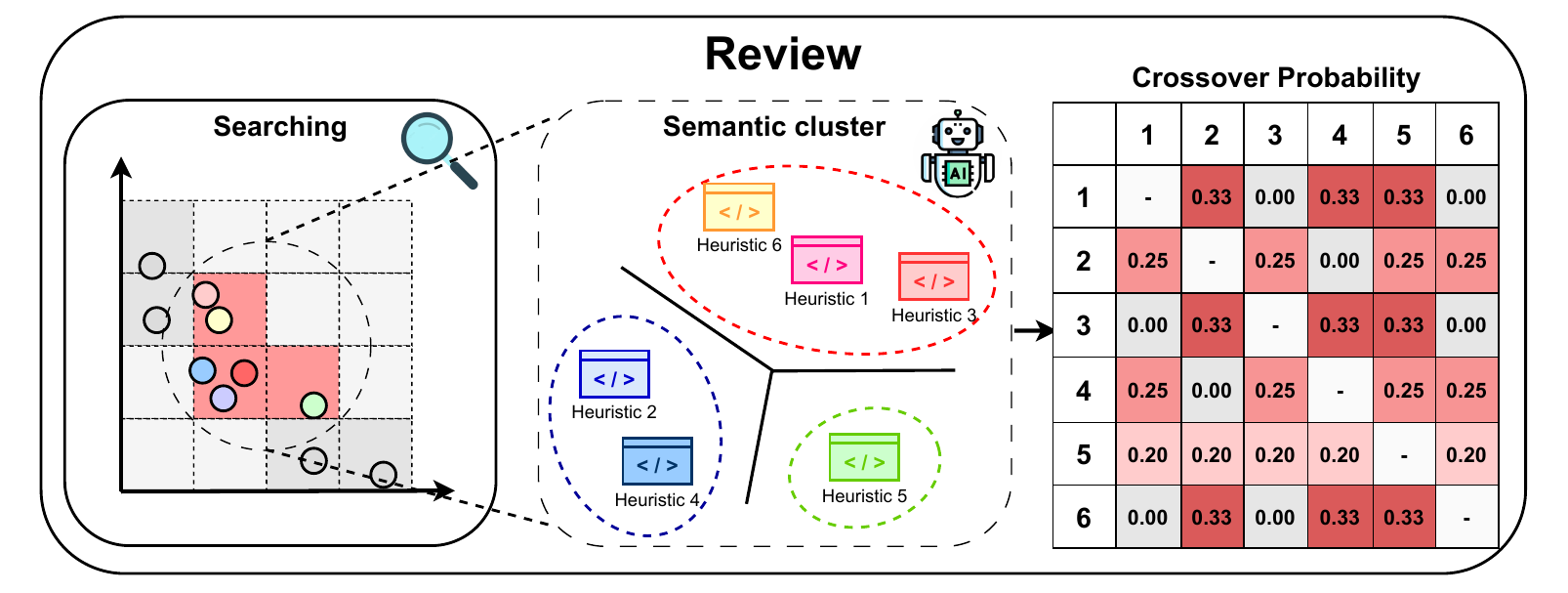}
    \caption{\small LLM-based review clustering applies elitism heuristics based on semantic similarity and employs a probability matrix to guide variation among clusters.}
    \label{fig:selectionpfg}
    \vspace{-1em}
\end{figure}
To address these challenges, instead of relying solely on syntactic features, we query \gls{llms} to assess the semantic and behavioral similarity among elite heuristics \( P = \{h_1, h_2, \ldots, h_n\} \), and group them into coherent clusters as depicted in Figure \ref{fig:selectionpfg}:
\begin{equation}
\begin{aligned}
\text{SemClust}(P) &= \{C_1, C_2, \ldots, C_m\} \\
\text{s.t.} \quad & \bigcup\nolimits_{i=1}^m C_i = P, \\
                 & C_i \cap C_j = \emptyset \quad (i \ne j).
\end{aligned}
\end{equation}
\noindent Each cluster \( C_i \) contains heuristics whose code segments implement similar underlying logic, despite syntactic or structural differences. This semantic clustering mitigates redundancy and enhances behavioral diversity at a higher abstraction level. During variation, we apply mutation within clusters to explore local variations, and perform crossover across clusters to combine diverse behaviors. 
Specifically, given a randomly selected cluster \( C_i \) and a heuristic \( h \in C_i \), the offspring heuristic \( o \) is generated as follows:
\begin{equation}
o \leftarrow 
\begin{cases}
\text{Mutate}(h), & \text{if } \mathcal{U}[0,1] < \gamma, \\
\text{Crossover}(h, h'), & \text{otherwise}, \quad h' \sim \mathcal{U}\left(\bigcup_{k \ne i} C_k\right),
\end{cases}%
\end{equation}
More detailed insights can be found in Appendix F.
\vspace{-0.5em}

\subsection{Automatic Heuristic generation}
\label{Automatic Heuristic generation}
In this section, we propose the \gls{mpage} framework for automatic heuristic design.
Examples of the prompts used at each stage are provided in the Appendix E. 

\textbf{Individual Representation.} Following previous works \citep{liu2024eoh, ye2024reevo}, \gls{mpage} encodes each individual as a natural language description and its corresponding code implementation in Python, both generated by \gls{llms}, together with an associated fitness score. The fitness is evaluated over a set of problem-specific instances by measuring both performance and running time.

\textbf{Population initialization.}
\gls{mpage} initializes the heuristic population by querying \gls{llms} with prompts that describe the problem and specify the signature of the heuristic function to be discovered. Specifically, each heuristic is tailored for selection mechanisms and neighborhood exploration within the \gls{semo} paradigm, then evaluated on benchmark instances and assigned a fitness score.

\textbf{Selection.} \gls{mpage} partitions the objective space into grid cells to construct an elitism pool (Section~\ref{Pareto Front Grid Generation}), then clusters the candidates and selects parent pairs based on dissimilarity as detailed in Section~\ref{Logical Assessment Mechanism}, or sample
from the entire population as Equation~\ref{eq:selection}.

\textbf{Feedback reflection.} Feedback reflection enhances heuristic design by providing \gls{llms} with clear, interpretable signals that guide improvement.
For each pair of heuristic parents, an LLM-based reflection module analyzes their respective strengths and weaknesses, returning a textual suggestion on how to improve or effectively combine them.

\textbf{Crossover and Mutation.}
\gls{mpage} prompts the generator LLM to create an offspring heuristic by recombining or modifying parent elements, guided by reflective feedback. The prompt includes: (i) task specifications, (ii) parent heuristics in code and natural language, (iii) reflection guidance, and (iv) instructions to generate the new heuristic.

\textbf{Population Management.} Offspring are merged into the population, with non-dominated individuals forming the next generation, as defined by Equation~\ref{eq:union}. The process repeats until the maximum number of iterations is reached.

\vspace{-1em}
\section{Experiments}

\subsection{Experimental Setup}

\subsubsection*{Benchmarks.} 
We evaluate the proposed \gls{mpage} framework on four widely recognized \gls{mocop} that are extensively investigated in the literature: Bi-TSP, Tri-TSP,  Bi-CVRP, and Bi-KP.
Comprehensive descriptions of these problems are provided in Appendix~A.
\subsubsection*{Experiment settings.} The generated heuristics are evaluated on 10 instances for each problem, with sizes of 20, 20, 50, and 50 for Bi-TSP, Tri-TSP, Bi-CVRP and Bi-KP, following the setup in \citep{chen2023efficient}.
The experimental parameters are configured as follows: the number of generations is set to 20, and the population size is 10 for all problems. Each crossover operator selects 2 parent heuristics to generate offspring heuristics. All heuristics are designed and evaluated under the SEMO search paradigm, with a runtime constraint of 2000 iterations and a time limit of 60 seconds. The probabilities \( \epsilon = 0.9 \), \( \gamma = 0.3 \), and the number of PFG segments is set to 4. Experiments were conducted on a Mac with an Apple M1 chip and 8\,GB RAM. \texttt{GPT-4o-mini} (temperature 0.7) was used as the pre-trained LLM, while \texttt{GPT-4o} was used for assessing and clustering task.

\subsubsection*{Performance Metric.} 
\subsubsection*{Objectives}
To evaluate a heuristic, we consider two objectives, both averaged across instances and minimized simultaneously: 1)  \textit{Negative Hypervolume (NHV)}: Minimizing the negative hypervolume guides the search toward high-quality solutions. 2) \textit{Running Time}: Measures the execution time of the heuristic to encourage efficiency.
\subsubsection*{Metrics} 
Solution quality is evaluated using the hypervolume (HV) and Inverted Generational Distance (IGD). HV indicates how well the Pareto front is approximated (higher is better), while IGD measures proximity and distribution relative to a reference front. Specifically, when comparing LLM-based methods, HV captures overall performance, as each heuristic is evaluated by NHV and execution time; for MOEAs, it reflects final solution quality.
In term of diversity measurement, we use the Shannon-Wiener Diversity Index (SWDI) and Cumulative Diversity Index (CDI) \citep{dat2025hsevo}. Details are provided in Appendix D.
\subsubsection*{Baseline Methods}  
For LLM-based automated heuristic design, we compare our method against representative approaches: EoH, MEoH, ReEvo and HSEvo \citep{liu2024eoh, yao2025meoh, ye2024reevo, dat2025hsevo} and additionally against widely adopted \gls{moeas}, including NSGA-II, MOEA/D, SEMO and PFG-MOEA \citep{deb2002fast, li2008multiobjective, xu2023pareto, laumanns2004runtime}. See Appendix B for the full experimental setup and descriptions.

\begin{figure*}[t]
    \centering

    \begin{subfigure}{0.40\textwidth}
        \centering
        \includegraphics[width=\linewidth,height=3.6cm]{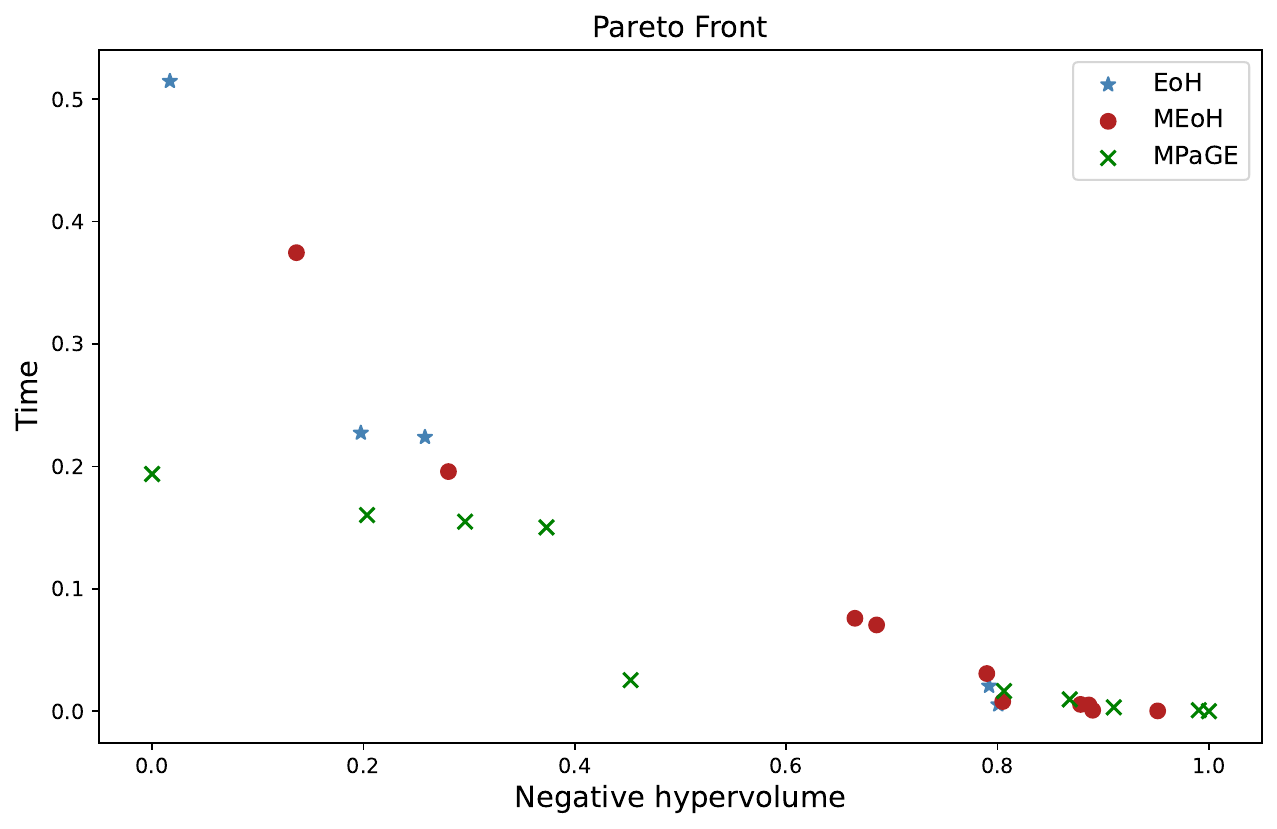}
        \caption*{\small Figure 3a: Pareto Front of heuristics on Bi-TSP20}
        \label{fig:a}
    \end{subfigure}
    \hfill
    \begin{subfigure}{0.58\textwidth}
     
        \centering
        \includegraphics[width=\linewidth,height=3.6cm]{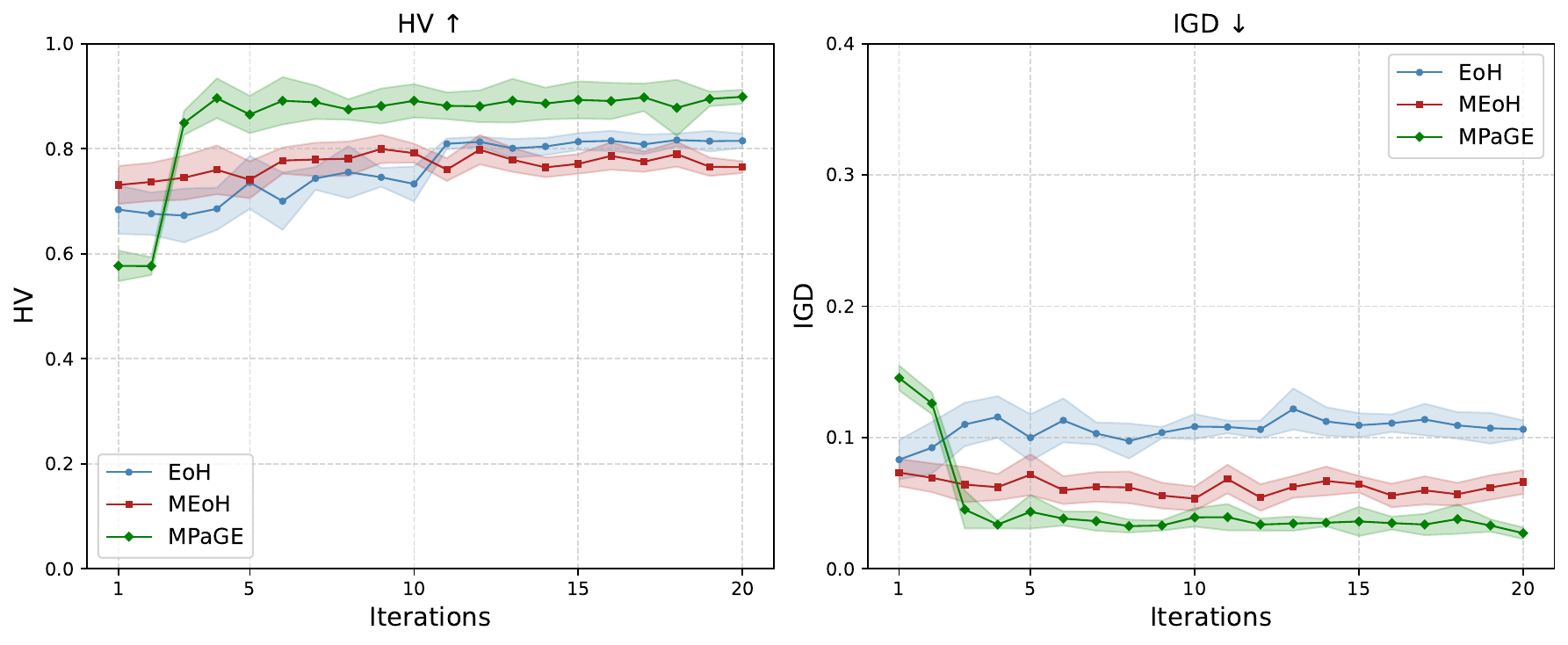}
        \caption*{\small Figure 3b: HV and IGD comparison on Bi-TSP20}\label{fig:b}
    \end{subfigure}

    \vspace{0em}

    \begin{subfigure}{0.40\textwidth}
        \centering
        \includegraphics[width=\linewidth,height=3.6cm]{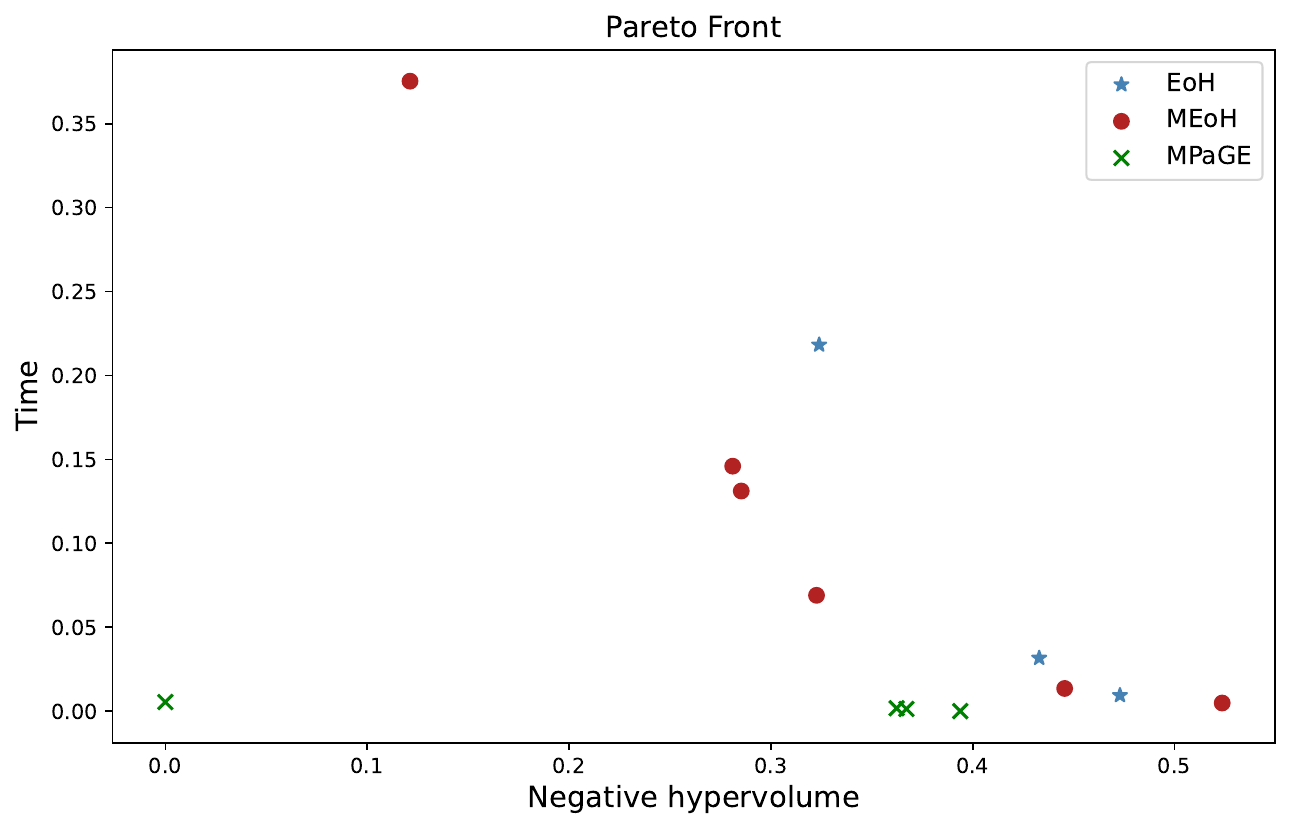}
        \caption*{\small Figure 3c: Pareto Front of heuristics  on Tri-TSP20}
        \label{fig:c}
    \end{subfigure}
    \hfill
    \begin{subfigure}{0.58\textwidth}
     
        \centering
        \includegraphics[width=\linewidth,height=3.6cm]{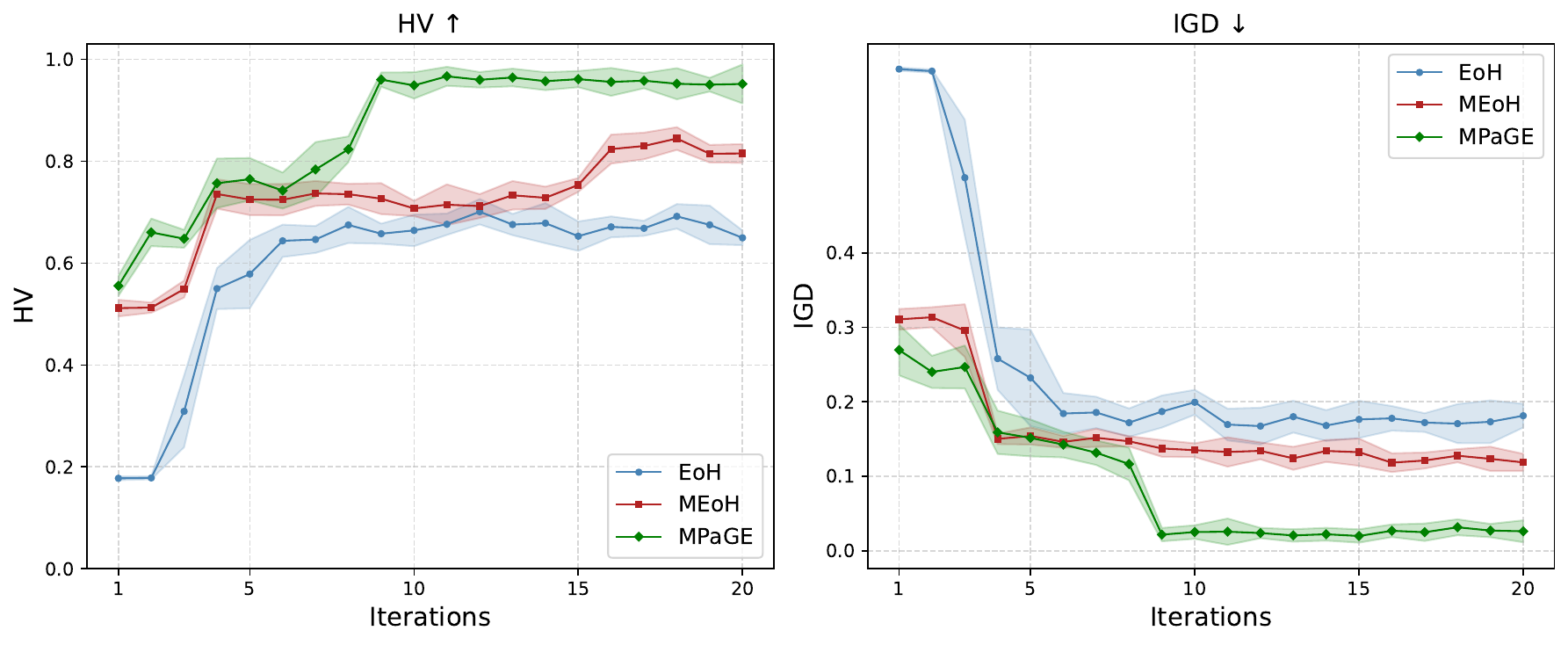}
        \caption*{\small Figure 3d: HV and IGD comparison on Tri-TSP20}
        \label{fig:d}
    \end{subfigure}

    \vspace{0em}

    \begin{subfigure}{0.40\textwidth}
        \centering
        \includegraphics[width=\linewidth,height=3.6cm]{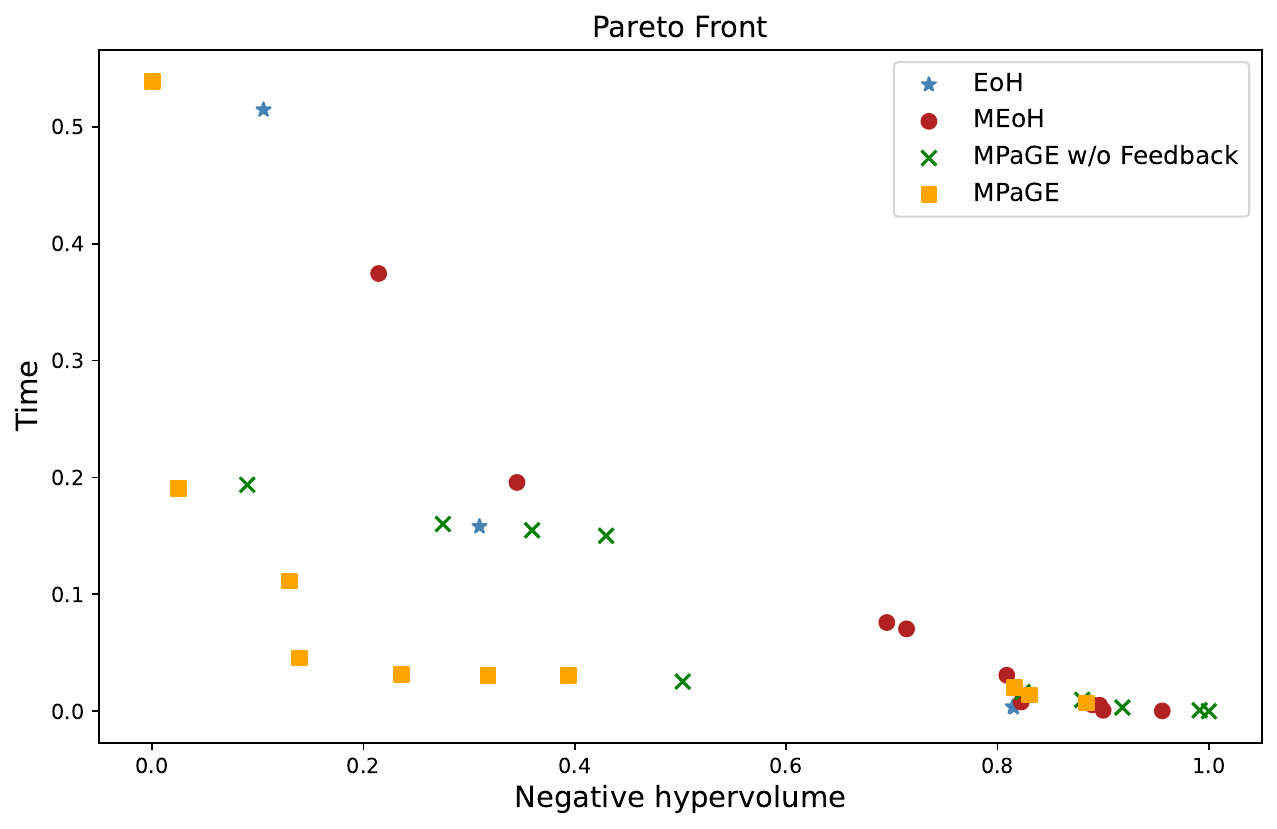}
        \caption*{\small Figure 3e: The non-dominated heuristics on Bi-TSP20}
        \label{fig:e}
    \end{subfigure}
    \hfill
    \begin{subfigure}{0.58\textwidth}
     
        \centering
        \includegraphics[width=\linewidth,height=3.6cm]{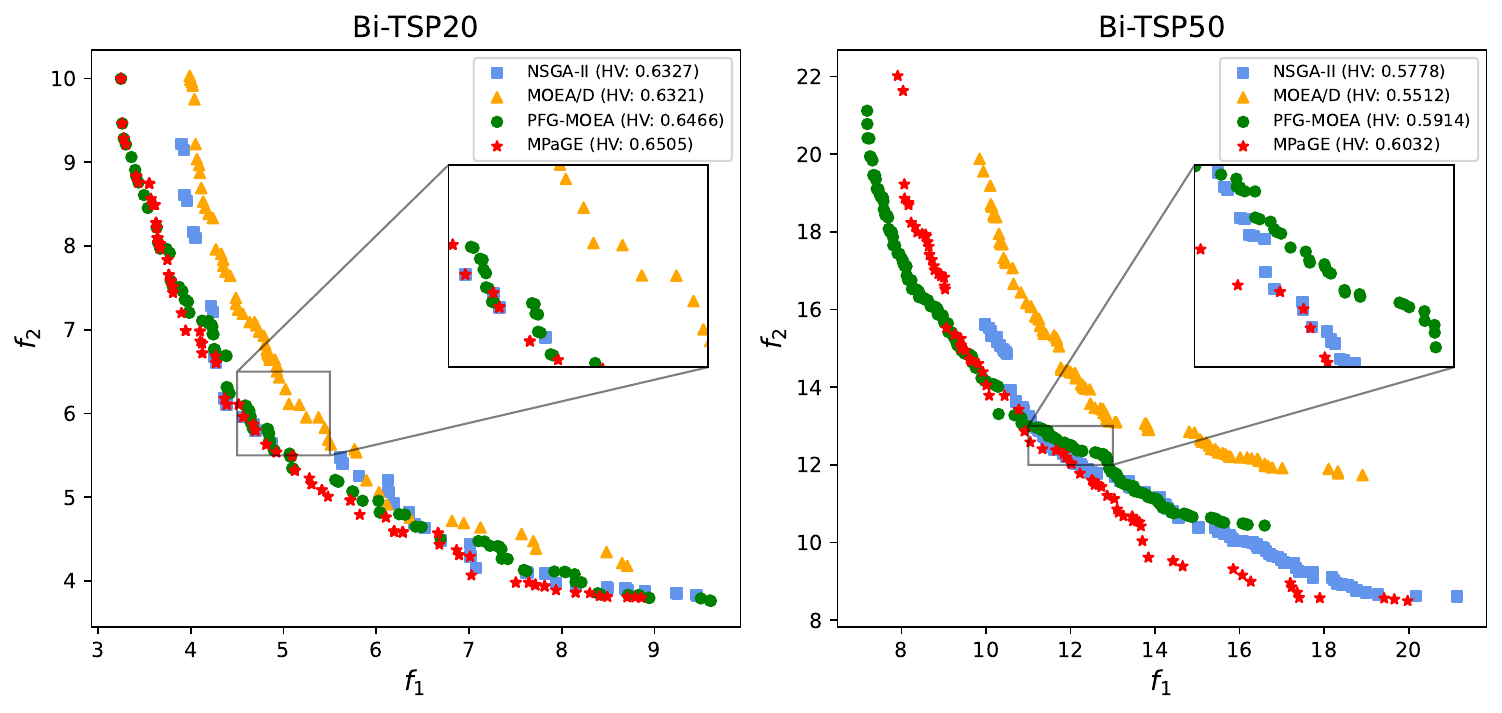}
        \caption*{\small Figure 3f: Pareto fronts of benchmark instances, Bi-TSP 20/50}\label{fig:f}
    \end{subfigure}

    \label{fig:all}
    \vspace{-1em}
\end{figure*}
\vspace{-1em}

\subsection{Experimental Results}

\subsubsection*{Pareto Fronts and Convergence Analysis} 
As illustrated in Table \ref{tab:table_1}, \gls{mpage} consistently outperforms other LLM-based baselines in both convergence and Pareto front quality across all benchmarks.
On Bi-TSP20 and Tri-TSP20, \gls{mpage} achieves the highest HV scores of 0.911 and 0.936, and the lowest IGD scores of 0.010 and 0.050, respectively, indicating faster convergence and superior solution quality. Similar trends are observed on Bi-KP50 and Bi-CVRP50, where \gls{mpage} maintains strong performance.
\vspace{-0.5em}
\begin{table}[!h]
\caption{\small Results of \gls{mpage} compared to other baselines regarding HV and IGD on four benchmark problems. The best values are highlighted in bold.}
\vspace{-1em}
\centering
\resizebox{\columnwidth}{!}{%
\begin{tabular}{lcccccccc}
\toprule
\multirow{2}{*}{Method} 
& \multicolumn{2}{c}{Bi-TSP20} 
& \multicolumn{2}{c}{Tri-TSP20} 
& \multicolumn{2}{c}{Bi-CVRP50} 
& \multicolumn{2}{c}{Bi-KP50} \\
\cmidrule(lr){2-3} \cmidrule(lr){4-5} \cmidrule(lr){6-7} \cmidrule(lr){8-9}
 & HV ↑ & IGD ↓ & HV ↑ & IGD ↓ & HV ↑ & IGD ↓ & HV ↑ & IGD ↓ \\
\midrule
EoH & 0.756 & 0.117 & 0.755 & 0.148 & 0.957 & 0.538 & 0.602 & 0.363 \\
ReEvo & 0.541 & 0.435 & 0.694 & 0.547 & 0.658 & 0.462 & \textbf{0.996} & 0.138 \\
HSEvo & 0.557 & 0.329 & 0.715 & 0.308 & 0.626 & 0.450 & 0.730 & 0.182 \\
MEoH & 0.724 & 0.067 & 0.884 & 0.114 & 0.322 & 0.286 & 0.748 & 0.248 \\
\gls{mpage} (Ours) & \textbf{0.911} & \textbf{0.010} & \textbf{0.936} & \textbf{0.050} & \textbf{0.980} & \textbf{0.007} & 0.932 & \textbf{0.035} \\
\bottomrule
\end{tabular}
}
\vspace{-1em}  
\label{tab:table_1}
\end{table}

In Figures 3a -- 3d, the non-dominated heuristics in the final population, along with the corresponding convergence curves of HV and IGD per iteration, illustrate clear differences among methods on Bi-TSP20 and Tri-TSP20, respectively. \gls{mpage} consistently yields a broader and more diverse set of Pareto-optimal solutions, spanning wider regions of the objective space, converges faster and clearly outperforms other baselines in terms of HV and IGD.  In contrast, EoH shows slower HV growth and higher IGD due to its narrow focus on minimizing performance gaps alone. MEoH, while enhancing diversity over EoH, still trails \gls{mpage} in convergence and final quality, likely due to the complexity of \gls{mocop} tasks and the limited impact of its diversity mechanism.
\vspace{-0.5em}

\begin{table}[!h]
\caption{\small Two top heuristics designed by MEoH and \gls{mpage}.}
\vspace{-1em}
\centering
\resizebox{\columnwidth}{!}{%
\begin{tabular}{lcccccccc}
\toprule
\multirow{2}{*}{Method} 
& \multicolumn{2}{c}{Bi-TSP20} 
& \multicolumn{2}{c}{Tri-TSP20} 
& \multicolumn{2}{c}{Bi-CVRP50} 
& \multicolumn{2}{c}{Bi-KP50} \\
\cmidrule(lr){2-3} \cmidrule(lr){4-5} \cmidrule(lr){6-7} \cmidrule(lr){8-9}
 & HV ↑ & Time ↓ & HV ↑ & Time ↓ & HV ↑ & Time ↓ & HV ↑ & Time ↓ \\
\midrule
MEoH (best) & 0.603 & 3.265 & 0.410 & 5.790 & 0.241 &  3.383  & 0.344 & 4.139 \\
MEoH (fast) & 0.345 & \cellcolor{gray!30}\textbf{0.177} & 0.403  & \cellcolor{gray!30}\textbf{5.228} & 0.141 & 0.081 & 0.349 & 2.246 \\
\gls{mpage} (best) & \cellcolor{gray!30}\textbf{0.629} & 4.429 & \cellcolor{gray!30}\textbf{0.478}  & 8.112 & \cellcolor{gray!30}\textbf{0.454}  & 0.191 & \cellcolor{gray!30}\textbf{0.359}  & 2.647 \\
\gls{mpage} (fast) & 0.451 & 1.304 & 0.411 & 7.590 & 0.151 & \cellcolor{gray!30}\textbf{0.063} & 0.354 &  \cellcolor{gray!30}\textbf{1.367} \\
\bottomrule
\end{tabular}
}
\vspace{-1em}  
\label{tab:table_2}
\end{table}

We evaluate the two best-so-far heuristics across multiple benchmark problems (Table~\ref{tab:table_2}), each in two configurations: ``best'' (highest hypervolume) and ``fast'' (lowest runtime), measured at the final population. Notably, \gls{mpage} discovers heuristics that outperform MEoH in HV. In several cases, its fast variant is quicker and more effective, highlighting \gls{mpage}'s strength in balancing quality and efficiency.
\vspace{-0.3em}

\subsubsection*{In and out-of-distribution size generalization analysis} 
\begin{table*}[ht]
\caption{\small Comparison results across all benchmarks against baselines. The HV and time are averaged over 50 instances.}
\vspace{-1em}
\centering
\resizebox{\textwidth}{!}{%
\scriptsize
\setlength{\tabcolsep}{3pt}
\begin{tabular}{l|
ccc|ccc|ccc|
ccc|ccc|ccc}
\toprule
\multirow{2}{*}{Method} 
& \multicolumn{3}{c|}{Bi-TSP20} & \multicolumn{3}{c|}{Bi-TSP50} & \multicolumn{3}{c|}{Bi-TSP100} 
& \multicolumn{3}{c|}{Tri-TSP20} & \multicolumn{3}{c|}{Tri-TSP50} & \multicolumn{3}{c}{Tri-TSP100} \\
& HV $\uparrow$ & Speedup & Time $\downarrow$& HV $\uparrow$ & Speedup & Time $\downarrow$ & HV $\uparrow$ & Speedup & Time $\downarrow$
& HV $\uparrow$ & Speedup & Time $\downarrow$ & HV $\uparrow$ $\uparrow$ & Speedup & Time $\downarrow$ & HV $\uparrow$ & Speedup & Time $\downarrow$ \\
\midrule
NSGA-II         & 0.603 & 1.0x & 731.2 & 0.495 & 1.0x & 875.5 & 0.419 & 1.0x & 932.5 & 0.451 & 3.9x & 252.446 & 0.279 & 1.7x & 669.9 & 0.195 & 1.5x & 830.3 \\
MOEA/D          & 0.597 & 5.8x & 126.644 & 0.473 & 5.5x & 157.796 & 0.414 & 4.0x & 240.680 & 0.365 & 4.5x & 216.485 & 0.189 & 3.6x & 310.583 & 0.123 & 3.1x & 398.387 \\
PFG-MOEA        & 0.624 & 1.2x & 617.734 & 0.532 & 1.0x & 844.914 & 0.457 & 1.0x & 952.186 & 0.465 & 1.0x & 984.180 & \cellcolor{gray!50}0.289 & 1.0x & 1127.46 & \cellcolor{gray!50}0.222 & 1.0x & 1229.857 \\
SEMO            & 0.543 & 149.4x & 4.894 & 0.284 & 80.5x & 10.878 & 0.178 & 56.5x & 16.846 & 0.295 & 143.0x & 6.881 & 0.108 & \textbf{84.8x} & \cellcolor{gray!50}13.290 & 0.065 & 33.2x & 36.994 \\
\midrule
EoH             & 0.584 & 52.2x & 14.004 & 0.510 & 18.5x & 47.233 & \cellcolor{gray!50}0.503 & 6.0x & 159.156 & 0.420 & 108.7x & 9.050 & 0.208 & 39.7x & 28.368 & 0.107 & 29.3x & 42.038 \\
ReEvo           & 0.623 & 107.3x & 6.812 & 0.516 & 86.2x & 10.151 & 0.413 & 51.0x & 18.653 & 0.427 & 95.0x & 10.361 & 0.205 & 41.1x & 27.416 & 0.103 & 14.8x & 83.213 \\
HSEvo           & 0.620 & 112.3x & 6.513 & 0.519 & 95.3x & 9.182 & 0.422 & 54.1x & 17.590 & 0.420 & 115.6x & 8.515 & 0.218 & 55.2x & 20.435 & 0.115 & 20.3x & 60.659 \\
MEoH (best)     & 0.603 & \textbf{224.0x} & \cellcolor{gray!50}3.265 & 0.480 & \textbf{126.2x} & \cellcolor{gray!50}6.937 & 0.395 & \textbf{79.2x} & \cellcolor{gray!50}12.024 & 0.410 & \textbf{170.0x} & \cellcolor{gray!50}5.790 & 0.205 & 79.4x & 14.197 & 0.111 & \textbf{48.8x} & \cellcolor{gray!50}25.213 \\
\gls{mpage} (best)   & \cellcolor{gray!50}0.629 & 165.1x & 4.429 & \cellcolor{gray!50}0.542 & 126.0x & 6.950 & 0.442 & 74.9x & 12.719 & \cellcolor{gray!50}0.478 & 121.3x & 8.112 & 0.220 & 76.5x & 14.747 & 0.109 & 46.6x & 26.394 \\
\midrule
\midrule
\multirow{2}{*}{Method} 
& \multicolumn{3}{c|}{Bi-KP50} & \multicolumn{3}{c|}{Bi-KP100} & \multicolumn{3}{c|}{Bi-KP200} 
& \multicolumn{3}{c|}{Bi-CVRP20} & \multicolumn{3}{c|}{Bi-CVRP50} & \multicolumn{3}{c}{Bi-CVRP100} \\
& HV $\uparrow$ & Speedup & Time $\downarrow$ & HV $\uparrow$& Speedup & Time $\downarrow$ & HV $\uparrow$ & Speedup & Time $\downarrow$
& HV $\uparrow$ & Speedup & Time $\downarrow$ & HV $\uparrow$ & Speedup & Time $\downarrow$ & HV $\uparrow$ & Speedup & Time $\downarrow$ \\
\midrule
NSGA-II         & 0.357 & 1.0$\times$ & 926.406 & 0.483 & 1.0$\times$ & 995.178 & 0.293 & 1.0$\times$ & 1176.915 & 0.573 & 1.0$\times$ & 1051.192 & 0.416 & 1.0$\times$ & 1945.667 & 0.357 & 1.0$\times$ & 3433.284 \\
MOEA/D          & 0.358 & 10.7$\times$ & 86.247 & 0.484 & 13.9$\times$ & 71.674 & 0.286 & 13.4$\times$ & 87.621 & 0.568 & 16.2x & 64.698 & 0.420 & 23.5x & 82.801 & 0.353 & 35.7x & 96.200 \\
PFG-MOEA        & 0.358 & 1.9x & 488.617 & 0.484 & 1.7x & 596.989 & \cellcolor{gray!50}0.330 & 2.1x & 570.174 & \cellcolor{gray!50}0.598 & 6.6x & 159.867 & 0.435 & 7.0x & 277.464 & 0.318 & 9.6x & 359.108 \\
SEMO            & 0.195 & 136.9x & 6.766 & 0.144 & 529.4x & 1.880 & 0.188 & 115.0x & 10.234 & 0.518 & 8342.8x & 0.126 & 0.205 & 1616.0x & 1.204 & 0.135 & 944.8x & 3.634 \\
\midrule
EoH             & 0.343 & 4.9x & 187.455 & 0.463 & 5.1x & 195.414 & 0.320 & 18.5x & 63.555 & 0.539 & 6332.5x & 0.166 & 0.443 & 6176.7x & 0.315 & 0.369 & 5731.7x & 0.599 \\
ReEvo           & 0.357 & \textbf{368.1x} & \cellcolor{gray!50}2.517 & 0.452 & 432.1x & 2.303 & 0.202 & 550.2x & 2.139 & 0.511 & 3822.5x & 0.275 & 0.354 & 2231.3x & 0.872 & 0.249 & 2257.3x & 1.521 \\
HSEvo           & 0.357 & 330.5x & 2.803 & 0.450 & 318.3x & 3.127 & 0.205 & 456.5x & 2.578 & 0.515 & 3185.4x & 0.330 & 0.405 & 1781.7x & 1.092 & 0.305 & 916.0x & 3.748 \\
MEoH (best)     & 0.344 & 223.8x & 4.139 & 0.464 & 219.1x & 4.543 & 0.321 & 275.3x & 4.275 & 0.503 & 2280.2x & 0.461 & 0.241 & 575.1x & 3.383 & 0.126 & 279.3x & 12.291 \\
\gls{mpage} (best)   & \cellcolor{gray!50}0.359 & 350.0x & 2.647 & \cellcolor{gray!50}0.486 & \textbf{535.3x} & \cellcolor{gray!50}1.859 & 0.197 & \textbf{555.7x} & \cellcolor{gray!50}2.118 & 0.568 & \textbf{14599.9x} & \cellcolor{gray!50}0.072 & \cellcolor{gray!50}0.454 & \textbf{10186.7x} & \cellcolor{gray!50}0.191 & \cellcolor{gray!50}0.422 & \textbf{9563.5x} & \cellcolor{gray!50}0.359 \\

\bottomrule
\end{tabular}
}
\label{comparsion}
\vspace{-1.5em}
\end{table*}
We evaluate the generalization capability of \gls{mpage} on both in-distribution and out-of-distribution larger instances of Bi-TSP (20, 50, 100, 150, 200), Tri-TSP (20, 50, 100), and Bi-KP instances (50, 100, 200), each comprising 10 instances. All hyperparameter settings are elaborated in Appendix D.
As evidenced in Table \ref{tab:table_3}, \gls{mpage} consistently outperforms baseline methods across various problem settings. While competing approaches suffer significant performance degradation as problem size increases, our method maintains remarkable stability and robustness. For instance, in Bi-TSP 200, \gls{mpage} achieves an IGD of just 0.017, substantially lower than MEoH (0.186) and EoH (0.181), indicating a much closer approximation to the true Pareto front. Similarly, in the more complex Tri-TSP 100, \gls{mpage} attains an IGD of 0.000, whereas MEoH and EoH produce 0.101 and 0.409, respectively, further demonstrating \gls{mpage}’s exceptional solution quality and scalability.

\vspace{-0.43em}

\subsubsection*{Impact of Reflection and Heuristic Diversity} 
We conducted experiments on the Bi-TSP20 benchmark to assess the effectiveness of our LLM Reflection approach. The results, summarized in Figure 3e and Table~\ref{tab:hv_igd}, demonstrate the superior performance of \gls{mpage} over baseline methods (EoH and MEoH) in both HV and IGD metrics. Remarkably, even without the reflection-based feedback, \gls{mpage} outperforms all baselines, indicating that it inherently benefits from capturing the correlation among objectives within each grid. When enhanced with the reflection feedbacks, \gls{mpage} exhibits further improvements, achieving the highest HV and the lowest IGD. These findings underscore the strength of our approach in promoting both convergence and diversity.
To investigate the impact of our approach on population diversity, we evaluate all frameworks using the SWDI and CDI metrics, as depicted in Table \ref{tab:table_5}. Notably, \gls{mpage} demonstrates significantly better performance compared to the other baselines. Higher SWDI and CDI values indicate more uniform heuristic distribution and greater population diversity, promoting effective exploration. Further analysis is provided in Appendix F.
\vspace{-0.5em}
\subsubsection*{Comparison to Conventional MOEAs}
We evaluate the impact of \gls{pfg} on the optimization process and compare its performance with two well-established \gls{moeas}: NSGA-II \cite{deb2002fast} and MOEA/D \cite{zhang2007moea}. As shown in Table~\ref{tab:moeas}, based on experiments conducted on Bi-TSP problems, \gls{mpage} consistently outperforms the baselines in HV and IGD. These results highlight the effectiveness of the PFG mechanism. By directing the search toward the most promising regions of the objective space, the method enhances both the solution quality and the overall efficiency of the optimization process.
\vspace{-0.5em}
\begin{table}[ht]
  \caption{\small (a) Effects of LLM Reflection Bi‐TSP20; (b) Comparison in Shannon–Wiener diversity index (SWDI) and cumulative diversity index (CDI).}
  \vspace{-1em}
  \centering
  \tiny
  \setlength{\tabcolsep}{2pt} 
  \begin{subtable}[t]{0.45\columnwidth}
    \centering
    \renewcommand{\arraystretch}{1.1} 
    \resizebox{\linewidth}{!}{%
      \begin{tabular}{@{}lcc@{}}
        \toprule
        \textbf{Method} & \textbf{HV $\uparrow$} & \textbf{IGD $\downarrow$} \\
        \midrule
        EoH                  & 0.688 & 0.141 \\
        MEoH                 & 0.659 & 0.122 \\
        \gls{mpage} w/o Feedback  & 0.829 & 0.077 \\
        \textbf{\gls{mpage}}      & \cellcolor{gray!30}\textbf{0.941} & \cellcolor{gray!30}\textbf{0.023} \\
        \bottomrule
      \end{tabular}%
    }
    \subcaption{}
    \label{tab:hv_igd}
  \end{subtable}%
  \hspace{0.05\columnwidth}%
  \begin{subtable}[t]{0.45\columnwidth}
    \centering
    \renewcommand{\arraystretch}{1.1}
    \resizebox{\linewidth}{!}{%
      \begin{tabular}{@{}l|cc|cc@{}}
        \toprule
        \textbf{Problems} 
          & \multicolumn{2}{c|}{\textbf{Bi TSP}} 
          & \multicolumn{2}{c}{\textbf{Bi CVRP}} \\ 
        \cmidrule(lr){2-3} \cmidrule(lr){4-5} 
        & SWDI $\uparrow$ & CDI $\uparrow$ & SWDI $\uparrow$ & CDI $\uparrow$ \\
        \midrule
        EoH    & 0.897 & 1.944 & 1.168 & 2.173 \\
        ReEvo  & 0.647 & 2.133 & 0.943 & 1.908 \\
        HSEvo  & 0.757 & 1.915 & 1.102 & 1.964 \\
        MEoH   & 0.639 & 2.086 & 0.143 & 2.181 \\
        \gls{mpage} & \cellcolor{gray!30}\textbf{1.029} & \cellcolor{gray!30}\textbf{2.152} & \cellcolor{gray!30}\textbf{1.172} & \cellcolor{gray!30}\textbf{2.213} \\
        \bottomrule
      \end{tabular}%
    }
    \subcaption{}
    \label{tab:table_5}
  \end{subtable}
  \label{tab:table_4_main}
  \vspace{-2em}
\end{table}
\vspace{-1em}
\begin{table}[ht]
\caption{\small Performance comparison in terms of HV and IGD on in- and out-of-distribution instances.}
\vspace{-0.5em}
\centering
\resizebox{\columnwidth}{!}{%
\begin{tabular}{l|cc|cc|cc}
\toprule
\multirow{2}{*}{Problems} & \multicolumn{2}{c|}{EoH} & \multicolumn{2}{c|}{MEoH} & \multicolumn{2}{c}{\gls{mpage}} \\ 
\cmidrule(lr){2-3} \cmidrule(lr){4-5} \cmidrule(lr){6-7}
& HV ↑ & IGD ↓ & HV ↑ & IGD ↓ & HV ↑ & IGD ↓ \\
\midrule
Bi-TSP 20   & 0.843 & 0.100 & 0.786 & \textbf{0.045} & \textbf{0.918} & 0.063 \\
Bi-TSP 50   & 0.293 & 0.154 & 0.450 & 0.113 & \textbf{0.972} & \textbf{0.020} \\
Bi-TSP 100  & 0.361 & 0.169 & 0.364 & 0.134 & \textbf{0.937} & \textbf{0.026} \\
Bi-TSP 150  & 0.351 & 0.216 & 0.364 & 0.159 & \textbf{0.985} & \textbf{0.026} \\
Bi-TSP 200 & 0.341 & 0.181 & 0.349 & 0.186 & \textbf{0.990} & \textbf{0.017} \\
\midrule
Tri-TSP 20   & 0.755 & 0.148 & 0.884 & 0.114 & \textbf{0.936} & \textbf{0.050} \\
Tri-TSP 50  & 0.483 & 0.379 & 0.881 & 0.093 & \textbf{0.916} & \textbf{0.000} \\
Tri-TSP 100  & 0.447 & 0.409 & 0.810 & 0.101 & \textbf{0.897} & \textbf{0.000} \\
\midrule
Bi-KP 50   & 0.695 & 0.470 & 0.862 & 0.061 & \textbf{0.907} & \textbf{0.008} \\
Bi-KP 100  & 0.751 & 0.390 & 0.828 & 0.243 & \textbf{0.923} & \textbf{0.063} \\
Bi-KP 200  & 0.316 & 0.704 & 0.472 & 0.192 & \textbf{0.855}& \textbf{0.080} \\
\bottomrule
\end{tabular}
} 
\vspace{-1.3em} 
\label{tab:table_3}
\end{table}
\vspace{-0.3em}
\begin{table}[ht]
\caption{\small Efficiency of our Pareto Front Grid}
\vspace{-1em}
\centering
\resizebox{\columnwidth}{!}{%
\begin{tabular}{l|cc|cc|cc}
\toprule
\multirow{2}{*}{Backbone} & \multicolumn{2}{c|}{Bi-TSP 20} & \multicolumn{2}{c|}{Bi-TSP 50} & \multicolumn{2}{c}{Bi-TSP 100} \\ 
\cmidrule(lr){2-3} \cmidrule(lr){4-5} \cmidrule(lr){6-7}
& HV ↑ & IGD ↓ & HV ↑ & IGD ↓ & HV ↑ & IGD ↓ \\
\midrule
NSGA-II   &0.860 & 0.052 & 0.801 & 0.120 & 0.757 & \textbf{0.095} \\
MOEA/D   &0.819 & 0.119 & 0.768 & 0.108 & 0.560 & 0.157 \\
PFG (Ours)  & \textbf{0.913} & \textbf{0.024} & \textbf{0.836} & \textbf{0.075} & \textbf{0.844} & 0.099 \\
\bottomrule
\end{tabular}
} 
\vspace{-1em}
\label{tab:moeas}
\end{table}
\subsubsection*{Evaluation Against Baselines}
We evaluate the best heuristics generated by \gls{mpage} based on hypervolume performance, against baselines on standard \gls{mocop} benchmarks. 
As shown in Table \ref{tab:moeas} and Figure 3f, \gls{mpage} consistently outperforms existing LLM-based heuristics, achieving the highest HV on 9 out of 12 test suites and up to 100$\times$ speedup. Compared to traditional MOEAs, it delivers comparable or better HV on over half of the problems while being up to 14,599$\times$ faster. Although MEoH also offers strong runtime, \gls{mpage} achieves a more balanced trade-off, maintaining high HV even on large instances. For example, on Bi-TSP100 and Bi-CVRP100, it reaches HV of 0.442 and 0.422 while being 46.6$\times$ and 9563.5$\times$ faster than NSGA-II. Overall, \gls{mpage} offers a robust set of heuristics balancing optimality and efficiency, making it highly suitable for large-scale combinatorial optimization.

\vspace{-1em}
\section{Conclusion}
In this paper, we propose \gls{mpage}, a novel LLM-guided framework for solving MOCOP that simultaneously discovers a Pareto front of heuristics balancing solution quality, runtime efficiency, and semantic diversity. Integrating LLMs with the SEMO paradigm and introducing the Pareto Front Grid, our approach efficiently partitions the objective space and steers heuristic evolution toward promising regions. By clustering heuristics based on semantic logic and promoting inter-group diversity, the framework ensures meaningful variation within the heuristic population. 
Empirical results show that \gls{mpage} consistently outperforms prior LLM-based approaches in achieving superior trade-offs across objectives, enhancing heuristic diversity, and reducing computational cost. Furthermore, it outperforms traditional algorithms in efficiency while maintaining comparable solution quality, demonstrating its potential as a scalable and generalizable approach for automated heuristic discovery.

\bibliography{aaai26}

\cleardoublepage

\newpage
\appendix
\onecolumn

\section*{Appendix: Table of Contents}
\vspace{1em}

This is Appendix for ``Pareto-Grid-Guided Large Language Models for Fast and High-Quality Heuristics Design in Multi-Objective Combinatorial Optimization''.
\begin{itemize}[leftmargin=*, label={}]
    \item \textbf{A \quad Detailed of MOCOP benchmarks} \dotfill~\pageref{sec:mocop}
    \begin{itemize}[leftmargin=*, label={}]
        \item A.1 \quad Multi-objective traveling salesman problem (MOTSP) \dotfill~\pageref{subsec:motsp}
        \item A.2 \quad Multi-objective capacitated vehicle routing problem (MOCVRP) \dotfill~\pageref{subsec:mocvrp}
        \item A.3 \quad Multi-objective knapsack problem (MOKP) \dotfill~\pageref{subsec:mokp}
    \end{itemize}

    \item \textbf{B \quad Baseline Descriptions} 
    \dotfill~\pageref{app:moeas}
    \begin{itemize}[leftmargin=*, label={}]
        \item B.1 \quad LLM-based Baselines \dotfill~\pageref{llm-based}
        \item B.2 \quad MOEAs Baselines \dotfill~\pageref{moeas-based}
        \item B.3 \quad Motivation for Using SEMO as the Underlying Framework \dotfill~\pageref{why-semo}
    \end{itemize}

    \item \textbf{C \quad Algorithm details} \dotfill~\pageref{sec:algorithm}

    \item \textbf{D \quad Metric descriptions} \dotfill~\pageref{metric}
    \begin{itemize}[leftmargin=*, label={}]
        \item D.1 \quad Hypervolume \dotfill~\pageref{d1}
        \item D.2 \quad Inverted Generational Distance \dotfill~\pageref{d2}
        \item D.3 \quad Shannon-Wiener diversity index \dotfill~\pageref{d3}
        \item D.4 \quad Cummmulative diversity index \dotfill~\pageref{d4}
    \end{itemize}

    \item \textbf{E \quad \gls{mpage} prompt details} \dotfill~\pageref{e}
    \begin{itemize}[leftmargin=*, label={}]
        \item E.1 \quad Task description prompt \dotfill~\pageref{e1}
        \item E.2 \quad Initialization prompt \dotfill~\pageref{e2}
        \item E.3 \quad Semantic clustering prompt \dotfill~\pageref{e3}
        \item E.4 \quad Feedback reflection prompt \dotfill~\pageref{e4}
        \item E.5 \quad Crossover prompt \dotfill~\pageref{e5}
    \end{itemize}

    \item \textbf{F \quad Clustering variations analysis} \dotfill~\pageref{cluster}

    \item \textbf{G \quad Designed Heuristics} \dotfill~\pageref{design}

    \item \textbf{H \quad Comparison to Neural Combinatorial Optimization} \dotfill~\pageref{NCO}

\end{itemize}
\clearpage

\section{Detailed of \gls{mocop} benchmarks}
\label{sec:mocop}
\subsection{Multi-objective traveling salesman problem (MOTSP)}
\label{subsec:motsp}

In the multi-objective traveling salesman problem (MOTSP) involving $n$ nodes and $M$ objectives, each node $i \in \{1, \ldots, n\}$ is represented by $M$ sets of 2-dimensional coordinates, one for each objective. For a given objective $m$, the Euclidean distance $c_{ij}^m$ between two nodes $i$ and $j$ is calculated based on their corresponding coordinates. The goal is to construct a tour $\pi$ that visits every node exactly once and simultaneously minimizes the total travel distance across all $M$ objectives. Formally, the objective vector to minimize is defined as:
\[
f(\pi) = \left(f_1(\pi), f_2(\pi), \ldots, f_M(\pi)\right),
\]
with each component given by:
\[
f_m(\pi) = \sum_{i=1}^{n-1} c_{\pi_i, \pi_{i+1}}^m + c_{\pi_n, \pi_1}^m, \quad \forall m \in \{1, \ldots, M\}.
\]
Problem instances are synthetically generated by sampling all coordinates independently and uniformly from the $[0,1]^{2M}$ hypercube.

\subsection{Multi-objective capacitated vehicle routing problem (MOCVRP)}
\label{subsec:mocvrp}

We focus on the bi-objective capacitated vehicle routing problem (Bi-CVRP), which considers a set of $n$ customer nodes and a single depot. Each node, including the depot, is assigned a location in 2D space, and every customer has a specific demand. A fleet of identical vehicles with uniform capacity is stationed at the depot and must complete routes that collectively serve all customers, returning to the depot afterward. Each vehicle must have sufficient remaining capacity to fulfill the demand of any customer it visits. The problem simultaneously optimizes two conflicting objectives: minimizing the total traveled distance and minimizing the makespan, defined as the length of the longest route. In our Bi-CVRP instances, node coordinates are randomly sampled from the $[0,1]^2$ space, and customer demands are randomly selected from the set $\{1, \ldots, 9\}$. The vehicle capacity is set to 30, 40, and 50 for problem sizes where $20 \leq n < 40$, $40 \leq n < 70$, and $70 \leq n \leq 100$, respectively. To standardize inputs, all demand values are normalized with respect to vehicle capacity.

\subsection{Multi-objective knapsack problem (MOKP)}
\label{subsec:mokp}

In the multi-objective knapsack problem (MOKP) with $M$ objectives and $n$ items, each item is characterized by a weight and $M$ distinct profit values, one for each objective. These items can be visualized as nodes within an instance graph. The goal is to select a subset of items such that the total profit across all $M$ objectives is maximized, while the combined weight of the selected items does not exceed a predefined capacity. Formally, let $x \in \{0,1\}^n$ be a binary decision vector, where $x_i = 1$ indicates that item $i$ is selected. The problem can be stated as:

\[
\text{maximize } f(x) = \left(f_1(x), f_2(x), \ldots, f_M(x)\right),
\]
subject to:
\[
\sum_{i=1}^{n} w_i x_i \leq C, \quad x_i \in \{0, 1\}, \quad \forall i \in \{1, \ldots, n\},
\]
where $w_i$ is the weight of item $i$, and $f_m(x) = \sum_{i=1}^{n} p_i^m x_i$ represents the total profit for objective $m$, with $p_i^m$ being the profit of item $i$ under objective $m$.

The instances are generated by independently sampling all weights $w_i$ and profits $p_i^m$ from the uniform distribution over $[0, 1]$. The knapsack capacity $C$ is set to 12.5 for $50 \leq n < 100$ and 25 for $100 \leq n \leq 200$, respectively.

\section{Baseline Descriptions}\label{app:moeas}
\subsection{LLM-based Baselines}\label{llm-based}
We compare the Pareto fronts of heuristics generated by \gls{mpage} with those of the following methods:
\begin{itemize}
    \item \textbf{EoH \citep{liu2024eoh}} is a framework that combines large language models and evolutionary computation to automatically design heuristics. It simulates expert reasoning by evolving both code and thought via five tailored prompt strategies.
    \item \textbf{ReEvo \citep{ye2024reevo}}  is an evolutionary framework that integrates LLMs with reflective feedback to enhance heuristic generation. 
    \item \textbf{HSEvo \citep{dat2025hsevo}} is a framework built on harmony search principles, introducing new components and enhancing evolutionary operators to jointly optimize objective performance and solution diversity.
    \item \textbf{MEoH \citep{yao2025meoh}} is an LLM-based framework that formulates heuristic design as a multi-objective optimization problem to generate diverse, non-dominated heuristics in a single run.
\end{itemize}
To ensure fairness, experimental parameters follow \gls{mpage}: 20 generations and a population size of 10 across all problems. Each crossover operator selects two parent heuristics to generate offspring.

\subsection{\gls{moeas} Baselines}\label{moeas-based}
For all algorithms, we assume a minimisation problem as:
\[
\min_{x\in\mathcal X}\;F(x)=\bigl(f_1(x),\ldots,f_M(x)\bigr),
\]  
where~$F:\mathcal X\rightarrow\mathbb R^{M}$ is vector-valued with mutually conflicting objectives, and Pareto-dominance $\preceq$ is defined in the usual way.
To benchmark the effectiveness of LLM-generated heuristics in solving multi-objective combinatorial optimization problems, we compare them against three well-established \gls{moeas} and \gls{semo}. These algorithms are representative of different design philosophies in evolutionary multi-objective optimization and are widely used in the literature.

\subsubsection*{NSGA-II \cite{deb2002fast}}

NSGA-II maintains a population $\mathcal P_t$ of fixed size~$N$. It contains components:

\begin{itemize}
    \item \textbf{Fast non-dominated sorting.}\;Each solution $x$ is assigned a \emph{rank}  
          \[
          r(x)=\min\{k \in \mathbb{N}_0 \mid x \in F_k\},
          \]
          where $F_k$ is the $k$-th non-dominated front. Fronts are constructed iteratively: $F_0$ contains all non-dominated individuals, $F_1$ those dominated only by members of $F_0$, and so on. The total sorting cost is $O(MN^2)$ (or $O(MN\log N)$ with incremental approaches).
    \item \textbf{Crowding distance.}\;Within each front $F_r$, crowding distance is calculated as  
          \[
          d(x)=\sum_{m=1}^{M}\frac{f_{m}(x^{\,\text{next}})-f_{m}(x^{\,\text{prev}})}
                                             {f_{m}^{\max}-f_{m}^{\min}},
          \]
          where $x^{\text{prev}}$ and $x^{\text{next}}$ are the nearest neighbours in front $F_r$ along each objective $f_m$, and missing values at boundaries are treated as zero.
    \item \textbf{Selection and variation.}\;Binary tournament based on the lexicographic key  
          $(r(x),-d(x))$ chooses parents; simulated binary crossover (SBX) and
          polynomial mutation create $N$ offspring.
    \item \textbf{Elitist replacement.}\;Parents and offspring are merged and the best~$N$
          solutions under $(r,-d)$ are kept, guaranteeing $O(1)$ elitism per generation.

The algorithm explicitly balances \emph{convergence} (via rank) and \emph{diversity} (via distance) while preserving the worst-case $O(N)$ memory footprint.
\end{itemize}

\subsubsection*{MOEA/D \cite{zhang2007moead}}

MOEA/D decomposes the multi-objective problem into $K$ single-objective \emph{sub-problems} using a set of weight vectors $\{\boldsymbol\lambda^{(k)}\}_{k=1}^{K}$ with
$\boldsymbol\lambda^{(k)}\in\Delta^{M-1}$, where
$\Delta^{M-1}$ is the unit simplex.

\begin{itemize}
    \item \textbf{Scalarisation.}
Typical aggregation functions include:

\[
\begin{aligned}
g_{\text{WS}}(x\mid\boldsymbol\lambda^{(k)}) &=
           \sum_{m=1}^{M}\lambda^{(k)}_m f_m(x),\\[4pt]
g_{\text{Tcheb}}(x\mid\boldsymbol\lambda^{(k)},\mathbf z^\ast) &=
           \max_{m}\,\lambda^{(k)}_m\,
           \bigl|f_m(x)-z_m^\ast\bigr|,\\[4pt]
g_{\text{PBI}}(x\mid\boldsymbol\lambda^{(k)},\mathbf z^\ast) &=
           d_1+\theta d_2\quad
           \begin{cases}
           d_1=\dfrac{\bigl(F(x)-\mathbf z^\ast\bigr)^{\!\top}\boldsymbol\lambda^{(k)}}
                         {\lVert\boldsymbol\lambda^{(k)}\rVert},\\
           d_2=\left\lVert F(x)-\bigl(\mathbf z^\ast+d_1\boldsymbol\lambda^{(k)}\bigr)\right\rVert,
           \end{cases}
\end{aligned}
\]

where $\mathbf z^\ast=(\min f_1,\dots,\min f_M)$ is the \emph{ideal} point.

\item \textbf{Neighbourhood collaboration.}
Each sub-problem~$k$ is allotted a neighbourhood
$\mathcal N_k$ comprising the $T$ closest weight vectors in Euclidean distance.
Variation operators are restricted to parents drawn from $\mathcal N_k$, and
offspring $x'$ may update the solutions of all neighbours $j\in\mathcal N_k$
if $g(x'\mid\boldsymbol\lambda^{(j)})<g(x^{(j)}\mid\boldsymbol\lambda^{(j)})$.

The decomposition transforms the $M$-objective search into parallelised scalar
searches of complexity $O(KT)$ per generation while implicitly promoting
solution diversity through the geometry of $\{\boldsymbol\lambda^{(k)}\}$.

\end{itemize}

\subsubsection*{PFG-MOEA \cite{xu2023}}

PFG-MOEA is a recent grid-guided extension of MOEA/D that introduces \gls{pfg}.

\begin{itemize}
    \item \textbf{Pareto-front normalisation.}
Let $\mathbf z^\ast$ and $\mathbf z^{\mathrm{nad}}$ be the ideal and nadir
approximations of the current population.
For any solution $x$ define the normalised objective vector
$\tilde{F}(x)=(f_1',\dots,f_M')$ with
$f_m'=\dfrac{f_m(x)-z_m^\ast}{z_m^{\mathrm{nad}}-z_m^\ast}$.

\item \textbf{Grid construction.}
Given a grid size $G\in\mathbb N$, the \emph{cell index} of $x$ is  
\[
\boldsymbol g(x)=\bigl(g_1,\dots,g_M\bigr),\qquad
g_m=\bigl\lfloor G\,f_m'\bigr\rfloor.
\]
Only the \emph{leading} (knee) solution in each occupied cell is retained in an
external \emph{PFG archive}, drastically lowering memory and update costs
to $O(|\text{PFG}|)$, with $|\text{PFG}|\ll N$ for moderate~$G$.

\item \textbf{Environmental selection.}
At every generation PFG-MOEA:
\begin{enumerate}[label=(\roman*)]
    \item updates $\mathbf z^{\mathrm{nad}}$ via a statistical estimator using
          a random sample of the population, and
    \item performs a \emph{grid-based knee-point selection}: within each
          $\boldsymbol g$ the solution maximising
          \(
            \kappa(x)=\sum_{m} \bigl|f_m'(x)-\tfrac12\bigr|
          \)
          is chosen to preserve extreme trade-offs.
\end{enumerate}


The algorithm then follows MOEA/D-style neighbourhood variation and update,
but the decisions are guided by the sparsified PFG archive, yielding superior
convergence/diversity on irregular fronts with an empirical time complexity
close to $O(KT+|\text{PFG}|)$ per generation.

\end{itemize}
\subsubsection*{SEMO \cite{li2024empirical}} is a minimalistic $(1+1)$ MOEA widely used in theoretical studies. At each step, it selects a random solution from the current archive and applies a simple mutation to generate one neighbor. The new solution is added to the archive if it is not dominated, while any dominated solutions are removed. Its key strength lies in its simplicity and efficiency in exploring the Pareto front through fast, randomized local search, making it especially effective on discrete and combinatorial problems.
\bigskip
\noindent

To ensure a fair and controlled comparison, all algorithms are configured with a population size of 300 and are evolved for 300 generations across all benchmark instances. This uniform experimental setup allows performance differences to be attributed primarily to the algorithmic design, rather than variations in computational budget or search effort. For the SEMO algorithm and the LLM-generated heuristics applied to SEMO, we run 20,000 iterations for Bi-TSP and Tri-TSP, and 10,000 iterations for Bi-KP and Bi-CVRP.

Combinatorial optimization problems are encoded appropriately based on their structure. For permutation-based problems such as Bi-TSP, Tri-TSP, and Bi-CVRP, we adopt standard permutation representations and apply combinatorial variation operators including Swap Mutation and Partially Mapped Crossover. For binary-encoded problems such as Bi-KP, we use bitstring representations and apply bit-flip mutation along with standard uniform crossover to maintain diversity and introduce variation during evolution.

\subsection{Motivation for Using SEMO as the Underlying Framework}
\label{why-semo}
In this work, we adopt the Simple Evolutionary Multi-objective Optimization (SEMO) framework as the basis for evaluating LLM-generated heuristics, motivated by both practical efficiency and algorithmic effectiveness:
\begin{itemize}
    \item \textbf{Lightweight and Computationally Efficient:} Evaluating LLM-generated heuristics involves running optimization routines repeatedly, often hundreds or thousands of times over combinatorial problems. Embedding each heuristic within full-fledged MOEAs such as NSGA-II or MOEA/D results in significant computational overhead. In contrast, SEMO follows a $(1+1)$ design that evaluates only one solution per iteration, which greatly reduces runtime and makes large-scale heuristic search tractable.
    \item \textbf{Surprisingly Effective Search Performance:} Despite its simplicity, SEMO has demonstrated competitive, and in some cases superior, performance compared to both traditional MOEAs and local search heuristics \citep{li2024empirical}. One key factor is its unbounded archive, which enables the generation of new solutions from the entire set of non-dominated individuals, in contrast to the fixed-size populations used in most MOEAs. This promotes greater diversity and adaptability in the search. Additionally, SEMO’s inherent stochasticity in selection and mutation introduces a less greedy search behavior, which helps the algorithm escape local optima more effectively than deterministic local search methods such as PLS or Anytime PLS \citep{dubois2015anytime, paquete2004pareto}.
\end{itemize}

Building on these advantages, we employ LLMs to design the \textit{selection} and \textit{neighborhood exploration} components within the SEMO framework, resulting in heuristics that combine speed with strong search performance for \gls{mocop}.
\clearpage
\section{Algorithm details}
\label{sec:algorithm}

In this section, we provide a detailed explanation of the \gls{pfg} generation process and the proposed \gls{mpage} Framework, as illustrated in Algorithm~\ref{alg:pfg_grid} and Algorithm~\ref{alg:llmpfg}, respectively.

\begin{algorithm}[H]
\caption{\small PFG Generation}
\label{alg:pfg_grid}
\setcounter{AlgoLine}{0}
\small

\KwIn{
    Population $\mathcal{H}$ of heuristics; \\
    Objective values $e_j(h)$ for $j = 1,2$; \\
    Number of grid segments $K_1, K_2$; \\
    Small positive value $\sigma$
}
\KwOut{Grid mapping $\mathcal{G} : \mathbb{N}^2 \to 2^{\mathcal{|H|}}$; Elite set $\mathcal{E}$}

\For{$j \gets 1$ \KwTo $2$}{
    $z^*_j \gets \min_{h \in \mathcal{H}} e_j(h)$ \tcp*{Ideal point of objective $j$}
    
    $z^n_j \gets \max_{h \in \mathcal{H}} e_j(h)$ \tcp*{Nadir point of objective $j$}
    
    $\delta_j \gets \frac{z^n_j - z^*_j + 2\sigma}{K_j}$ \tcp*{Cell width with margin}
}

\ForEach{$h \in \mathcal{H}$}{
    \For{$j \gets 1$ \KwTo $2$}{
        $g_j \gets \left\lfloor \frac{e_j(h) - z^*_j + \sigma}{\delta_j} \right\rfloor$ \tcp*{Grid index}
    }
    
    $G(h) \gets (g_1, g_2)$ \tcp*{Grid cell index for $h$}
    
    $\mathcal{G}(G(h)) \gets \mathcal{G}(G(h)) \cup \{h\}$ \tcp*{Assign $h$ to cell}
}

$\mathcal{E} \gets \emptyset$\;

\ForEach{cell $g$ in $\mathcal{G}$}{
    $\mathcal{C} \gets \mathcal{G}(g)$ \tcp*{Current set of solutions in cell}
    
    $\mathcal{G}(g) \gets \emptyset$ \tcp*{Reset cell content}
    
    \ForEach{$h \in \mathcal{C}$}{
        \If{$\nexists h' \in \mathcal{C},\ h' \neq h \wedge e(h') \prec e(h)$}{
            $\mathcal{G}(g) \gets \mathcal{G}(g) \cup \{h\}$ \tcp*{Keep non-dominated $h$}
        }
    }
}
$\mathcal{E} \gets \bigcup_{g \in \operatorname{dom}(\mathcal{G})} \mathcal{G}(g)$ \tcp*{Collect elite set from all non-empty cells}

\Return $\mathcal{G},\ \mathcal{E}$\;
\end{algorithm}

\begin{algorithm}[H]
\caption{\small \gls{mpage} Framework}
\label{alg:llmpfg}
\KwIn{
    Population size $N$; Iteration count $T$; \\
    Problem description $\Pi$; Pretrained LLM $\mathcal{L}$; \\
    Grid size $(K_1, K_2)$; Margin $\sigma$; \\
    Probabilities $\epsilon$ (local selection), $\gamma$ (mutation)
}
\KwOut{Final heuristic population $P^*$}
\setcounter{AlgoLine}{0}
$P_0 \gets \emptyset$\;

\For{$i \gets 1$ \KwTo $N$}{
    $o \gets$ \texttt{LLMGenerate}($\mathcal{L}, \Pi$)\;
    
    $P_0 \gets P_0 \cup \{o\}$\;
}

\For{$t \gets 1$ \KwTo $T$}{
    $\mathcal{G}, \mathcal{E} \gets$ \textbf{PFGGeneration}($P_{t-1}, e_1, e_2, K_1, K_2, \sigma$)\;

    \For{$i \gets 1$ \KwTo $N$}{
        $u \sim \mathcal{U}(0,1)$\;

        \uIf{$u < \epsilon$}{
            $g \sim \mathcal{U}(1, |\mathcal{G}|)$; 
            $\mathcal{N}(g) \gets$ \texttt{Neighbors}($g$); 
            $\mathcal{P} \gets \bigcup_{g' \in \{g\} \cup \mathcal{N}(g)} \mathcal{G}(g')$ \tcp*{Exploration}
            $\{C_1, \ldots, C_m\} \gets$ \texttt{SemClust}($\mathcal{P}; \mathcal{L}$)\;

            $i \sim \mathcal{U}(1, m)$; \quad $h \sim C_i$\;

            $v \sim \mathcal{U}(0,1)$\;
            
            \uIf{$v < \gamma$}{
                $P_{\text{parent}} \gets \{h\}$ \tcp*{Mutation}
            }
            \Else{
                $h' \sim \bigcup_{k \ne i} C_k$\;
                
                $P_{\text{parent}} \gets \{h, h'\}$ \tcp*{Crossover}
            }
        }

        \Else{
            $P_{\text{parent}} \sim \texttt{Sample}(\mathcal{E}, 2)$ \tcp*{Exploitation}
        }

        $\phi \gets$ \texttt{ReflectiveFeedback}($\mathcal{L}, P_{\text{parent}}$)\;

        $o \gets$ \texttt{SearchOffspring}($\mathcal{L}, \phi$)\;

        $P_{t-1} \gets P_{t-1} \cup \{o\}$\;
    }
}

$P^* \gets$ \texttt{NonDominatedSet}($P_{T}$)\;

\Return $P^*$
\end{algorithm}

\clearpage
\section{Metric descriptions}
\label{metric}

\subsection{Hypervolume}
\label{d1}

The hypervolume (HV) metric is a prevalent indicator used to assess the quality of solutions generated by multi-objective combinatorial optimization (MOCO) algorithms. It evaluates both convergence toward the Pareto front and the diversity of solutions, without relying on a known ground truth. Given a reference point $\mathbf{r} \in \mathbb{R}^M$, the hypervolume of a Pareto front $\mathcal{F}$ is denoted as $\mathrm{HV}_{\mathbf{r}}(\mathcal{F})$ and defined by:
\begin{equation}
    \mathrm{HV}_{\mathbf{r}}(\mathcal{F}) = \mu \left( \bigcup_{\mathbf{f}(\pi) \in \mathcal{F}} [\mathbf{f}(\pi), \mathbf{r}] \right),
\end{equation}
where $\mu$ is the Lebesgue measure, and $[\mathbf{f}(\pi), \mathbf{r}]$ represents the axis-aligned hyperrectangle spanning from the point $\mathbf{f}(\pi)$ to the reference point $\mathbf{r}$ in $M$ dimensions, i.e., $[f(\pi), r] = [f_1(\pi), r_1] \times \cdots \times [f_M(\pi), r_M]$.

To ensure fair comparison of HV values across different objectives, we normalize each objective value based on global approximations. Specifically, to compare the quality of the Pareto fronts produced by LLM-based method, we compute the ideal point $\mathbf{z}^{\text{ideal}} = (z^{\text{ideal}}_1, \ldots, z^{\text{ideal}}_M)^\top$ and nadir point $\mathbf{z}^{\text{nadir}} = (z^{\text{nadir}}_1, \ldots, z^{\text{nadir}}_M)^\top$ from the union of all approximated Pareto fronts $\mathcal{P}$ obtained by all heuristics:

\begin{equation}
    f'_i(\mathbf{x}) = \frac{f_i(\mathbf{x}) - z^{\text{ideal}}_i}{z^{\text{nadir}}_i - z^{\text{ideal}}_i},
\end{equation}

\noindent where
\(
z^{\text{ideal}}_i = \min\{v_i \mid \mathbf{v} \in \mathcal{P} \}, \quad
z^{\text{nadir}}_i = \max\{v_i \mid \mathbf{v} \in \mathcal{P} \}, \quad \forall i \in \{1, \ldots, M\}.
\)
This normalization maps all objective values to $[0, 1]$. The HV reference point is set to $\mathbf{r}^* = (1.1, \ldots, 1.1)^\top$.

To evaluate HV of a heuristic for given \gls{mocop},
The HV is normalized as $\text{HV}^{\prime}_{\mathbf{r}}(\mathcal{F}) = \text{HV}_{\mathbf{r}}(\mathcal{F}) / \prod_{i=1}^{M} |r_i - z_i|$, where $\mathbf{z}$ is an ideal point such that $z_i < \min\{f_i(\pi) \mid f(\pi) \in \mathcal{F} \}$ (or $z_i > \max\{f_i(\pi) \mid f(\pi) \in \mathcal{F} \}$ for maximization), $\forall i \in \{1, \ldots, M\}$. The  $\mathbf{r}$ and $\mathbf{z}$ are used across all methods for a given MOCOP, as summarized in Table~\ref{tab:ref_points}.
\begin{table}[h]
\centering
\caption{\small Reference points and ideal points for the MOCO problems.}
\label{tab:ref_points}
\begin{tabular}{llcc}
\toprule
\textbf{Problem} & \textbf{Size} & $\mathbf{r}$ & $\mathbf{z}$ \\
\midrule
\multirow{5}{*}{Bi-TSP} 
  & 20  & (20, 20)     & (0, 0) \\
  & 50  & (35, 35)     & (0, 0) \\
  & 100 & (65, 65)     & (0, 0) \\
  & 150 & (85, 85)     & (0, 0) \\
  & 200 & (115, 115)   & (0, 0) \\
\midrule
\multirow{3}{*}{Bi-CVRP}
  & 20  & (30, 8)      & (0, 0) \\
  & 50  & (45, 8)      & (0, 0) \\
  & 100 & (80, 8)      & (0, 0) \\
\midrule
\multirow{3}{*}{Bi-KP}
  & 50  & (5, 5)       & (30, 30) \\
  & 100 & (20, 20)     & (50, 50) \\
  & 200 & (30, 30)     & (75, 75) \\
\midrule
\multirow{3}{*}{Tri-TSP}
  & 20  & (20, 20, 20) & (0, 0) \\
  & 50  & (35, 35, 35) & (0, 0) \\
  & 100 & (65, 65, 65) & (0, 0) \\
\bottomrule
\end{tabular}
\end{table}




\subsection{Inverted Generational Distance}
\label{d2}
The Inverted Generational Distance (IGD) is a commonly used metric to assess the quality of solutions generated by multi-objective optimization methods. It captures both convergence to the Pareto front and diversity among solutions by averaging the shortest distances from each point in a reference set of Pareto-optimal solutions to the closest solution in the approximated front.
Given a reference front $Q$ and a non-dominated set $\mathcal{P}$ obtained by the algorithm, IGD is computed as:

\begin{equation}
\label{eq:igd}
\mathrm{IGD} = \frac{1}{|Q|} \sum_{q \in Q} \min_{p \in \mathcal{P}} \|q - p\|_2,
\end{equation}

\noindent where $\|\cdot\|_2$ denotes the Euclidean norm. Lower IGD values indicate that the solution set $\mathcal{P}$ is closer to the true Pareto front $Q$ in terms of both convergence and diversity. In this study, the reference set $Q$ is constructed as the non-dominated front derived from the union of all heuristics \citep{coello2004study}.

\subsection{Shannon-Wiener diversity index}
\label{d3}
The Shannon–Wiener Diversity Index (SWDI) provides a quantitative measure of population diversity at a given time step, based on the distribution of individuals into clusters \citep{dat2025hsevo}. In the context of heuristic search algorithms, this index reflects how evenly the population is spread across the search space.
Given a set of encoded individuals $\mathcal{V} = \{ \mathbf{v}_1, \dots, \mathbf{v}_n \}$, each individual is assigned to a cluster using cosine similarity. Let $C_i$ denote the $i$-th cluster, and let $M$ be the total number of individuals in all clusters. The proportion of individuals in cluster $C_i$ is defined as:
\[
p_i = \frac{|C_i|}{M}.
\]
The diversity score is then computed using the Shannon entropy:
\[
H(X) = -\sum_{i=1}^{N} p_i \log(p_i),
\]
where $N$ is the total number of clusters. A higher value of $H(X)$ indicates a more uniform distribution of individuals across clusters, suggesting better exploration of the search space. In contrast, a lower value implies that individuals are concentrated in fewer clusters, which may facilitate exploitation of promising regions but also increases the risk of premature convergence. In this study, the clustering procedure and associated hyper-parameters follow the configuration described in \citep{dat2025hsevo}.

\subsection{Cummmulative diversity index}
\label{d4}
In the context of heuristic search, the Cumulative Diversity Index (CDI) quantifies how well the diversity, or system energy, is distributed from a centralized state to a more dispersed configuration.\citep{dat2025hsevo}.

Let $A = \{ \mathbf{v}_1, \dots, \mathbf{v}_n \}$ be the set of all individuals in the archive, where each individual is represented by an embedding vector in a continuous space. To assess the diversity within $A$, a Minimum Spanning Tree (MST) is constructed over the set using Euclidean distances between individual vectors. The MST connects all individuals with a subset of $|A|-1$ edges such that the total edge length is minimized and no cycles are formed.
Let $d_i$ denote the length of the $i$-th edge in the MST. The probability associated with each edge is computed as:
\[
p_i = \frac{d_i}{\sum_{j=1}^{|A|-1} d_j},
\]
and the cumulative diversity is then defined using Shannon entropy:
\[
H(X) = -\sum_{i=1}^{|A| - 1} p_i \log(p_i),
\]
where $|A|$ is the archive size. Higher CDI values indicate a more distributed and diverse population, which is essential for maintaining a robust search process.

\clearpage
\section{\gls{mpage} prompt details}
\label{e}

\subsection{Task description prompt}
\label{e1}
Our goal is to design a heuristic function for generating high-quality neighbor solutions in specific \gls{mocop}. The task description provided in the prompt and the template of the Python code snippet is outlined below. The inputs include a solution archive and relevant problem-specific data, and the output should be a feasible neighbor solution.

\definecolor{mygreen}{rgb}{0,0.5,0}
\definecolor{myblue}{rgb}{0.1,0.1,0.5}
\definecolor{myred}{rgb}{0.6,0.1,0.1}
\definecolor{codebg}{rgb}{0.97,0.97,0.97}

\lstdefinelanguage{CustomPython}{
  language=Python,
  morekeywords={import,from,def,return},
  keywordstyle=\color{mygreen}\bfseries,
  commentstyle=\color{myred}\itshape,
  stringstyle=\color{purple},
  basicstyle=\ttfamily\scriptsize,
  backgroundcolor=\color{codebg},
  showstringspaces=false,
  breaklines=true
}

\begin{tcolorbox}[
    colframe=black!75!blue,
    colback=white,
    sharp corners=southwest,
    title=\textbf{Bi-TSP heuristic design task description and template program.},
    title style={font=\bfseries\color{myblue}},
    boxrule=0.6pt,
    arc=6pt,
    left=1mm, right=1mm,
    top=0.5mm, bottom=0.5mm
]

\textbf{\textcolor{myblue}{Task Description:}} 
You are solving a Bi-objective Travelling Salesman Problem (bi-TSP), where each node has two different 2D coordinates: 
$(x_1, y_1)$ and $(x_2, y_2)$, representing its position in two objective spaces. The goal is to find a tour visiting each node exactly once and returning 
to the starting node, while minimizing two objectives simultaneously: the total tour length in each coordinate space.

Given an archive of solutions, where each solution is a numpy array representing a TSP tour, and its corresponding objective 
is a tuple of two values (cost in each space), design a heuristic function named \texttt{select\_neighbor} that selects one solution from the archive 
and applies a novel or hybrid local search operator to generate a neighbor solution from it.

Please perform an intelligent random selection from among the solutions that show promising potential for further local improvement. 
Using a creative local search strategy that you design yourself, go beyond standard approaches to design a method that yields higher-quality 
solutions across multiple objectives. The function should return the new neighbor solution.

\vspace{1em}
\textbf{\textcolor{myblue}{Template Program:}}

\begin{lstlisting}[language=CustomPython]
import numpy as np
from typing import List, Tuple
import random 

def select_neighbor(
    archive: List[Tuple[np.ndarray, Tuple[float, float]]],
    instance: np.ndarray,
    distance_matrix_1: np.ndarray,
    distance_matrix_2: np.ndarray
) -> np.ndarray:
    """
    Select a promising solution from the archive and generate a neighbor solution from it.

    Args:
    archive: List of (solution, objective) pairs. Each solution is a numpy array of node IDs.
             Each objective is a tuple of two float values (cost in each space).
    instance: Numpy array of shape (N, 4). Each row contains coordinates in 2D spaces: (x1, y1, x2, y2).
    distance_matrix_1: Distance matrix in the first objective space.
    distance_matrix_2: Distance matrix in the second objective space.

    Returns:
    A new neighbor solution (numpy array).
    """
    base_solution = archive[0][0].copy()
    new_solution = base_solution.copy()
    new_solution[0], new_solution[1] = new_solution[1], new_solution[0]

    return new_solution
\end{lstlisting}

\end{tcolorbox}

\begin{tcolorbox}[
    colframe=black!75!blue,
    colback=white,
    sharp corners=southwest,
    title=\textbf{Tri-TSP heuristic design task description and template program.},
    title style={font=\bfseries\color{myblue}},
    boxrule=0.6pt,
    arc=6pt,
    left=1mm, right=1mm,
    top=0.5mm, bottom=0.5mm
]

\textbf{\textcolor{myblue}{Task Description:}} 
You are solving a Tri-objective Travelling Salesman Problem (tri-TSP), where each node has three different 2D coordinates: 
$(x_1, y_1)$, $(x_2, y_2)$, and $(x_3, y_3)$, representing its position in three objective spaces. 
The goal is to find a tour visiting each node exactly once and returning to the starting node, while minimizing three objectives simultaneously: 
the total tour length in each coordinate space.

Given an archive of non-dominated solutions, where each solution is a numpy array representing a TSP tour, 
and its corresponding objective is a tuple of three values (cost in each space), design a heuristic function named \texttt{select\_neighbor} 
that selects one solution from the archive and applies a novel or hybrid local search operator to generate a neighbor solution from it.
Please perform an intelligent random selection from among the solutions that show promising potential for further local improvement. 
Using a creative local search strategy of your own design, specifically tailored to effectively optimize across three objectives, 
go beyond standard approaches to design a method that yields higher-quality solutions across multiple objectives. 
The function should return the new neighbor solution.

\vspace{1em}
\textbf{\textcolor{myblue}{Template Program:}}

\begin{lstlisting}[language=CustomPython]
import numpy as np
from typing import List, Tuple
import random 

def select_neighbor(
    archive: List[Tuple[np.ndarray, Tuple[float, float, float]]],
    instance: np.ndarray,
    distance_matrix_1: np.ndarray,
    distance_matrix_2: np.ndarray,
    distance_matrix_3: np.ndarray
) -> np.ndarray:
    """
    Select a promising solution from the archive and generate a neighbor solution from it.

    Args:
    archive: List of (solution, objective) pairs. Each solution is a numpy array of node IDs.
             Each objective is a tuple of three float values (costs in each space).
    instance: Numpy array of shape (N, 6). Each row contains coordinates: (x1, y1, x2, y2, x3, y3).
    distance_matrix_1: Distance matrix in the first objective space.
    distance_matrix_2: Distance matrix in the second objective space.
    distance_matrix_3: Distance matrix in the third objective space.

    Returns:
    A new neighbor solution (numpy array).
    """
    base_solution = archive[0][0].copy()
    new_solution = base_solution.copy()
    new_solution[0], new_solution[1] = new_solution[1], new_solution[0]

    return new_solution
\end{lstlisting}

\end{tcolorbox}

\begin{tcolorbox}[
    colframe=black!75!blue,
    colback=white,
    sharp corners=southwest,
    title=\textbf{Bi-KP heuristic design task description and template program.},
    title style={font=\bfseries\color{myblue}},
    boxrule=0.6pt,
    arc=6pt,
    left=1mm, right=1mm,
    top=0.5mm, bottom=0.5mm
]

\textbf{\textcolor{myblue}{Task Description:}} 
You are solving a Bi-objective Knapsack Problem (BI-KP), where each item has a weight and two profit values: 
\texttt{value1} and \texttt{value2}. The goal is to select a subset of items such that the total weight does not exceed a given capacity, 
while simultaneously maximizing the total value in both objective spaces.

Given an archive of non-dominated solutions, where each solution is a binary numpy array indicating item inclusion (1) or exclusion (0), 
and its corresponding objective is a tuple of two values (total \texttt{value1}, total \texttt{value2}), 
design a heuristic function named \texttt{select\_neighbor} that selects one solution from the archive 
and applies a novel or hybrid local search operator to generate a neighbor solution from it.

You must ensure that the generated neighbor solution remains feasible.
Please perform an intelligent random selection from among the solutions that show promising potential for further local improvement. 
Using a creative local search strategy that you design yourself, go beyond standard approaches to develop a method 
that yields higher-quality solutions across multiple objectives. 
The function should return the new neighbor solution.

\vspace{1em}
\textbf{\textcolor{myblue}{Template Program:}}

\begin{lstlisting}[language=CustomPython]
import numpy as np
from typing import List, Tuple
import random 

def select_neighbor(
    archive: List[Tuple[np.ndarray, Tuple[float, float]]],
    weight_lst: np.ndarray,
    value1_lst: np.ndarray,
    value2_lst: np.ndarray,
    capacity: float 
) -> np.ndarray:
    """
    Select a promising solution from the archive and generate a neighbor solution from it.

    Args:
    archive: List of (solution, objective) pairs. Each solution is a binary numpy array (0/1) of item selections.
             Each objective is a tuple of two float values (total value1, total value2).
    weight_lst: Numpy array of shape (N,), item weights.
    value1_lst: Numpy array of shape (N,), item values for objective 1.
    value2_lst: Numpy array of shape (N,), item values for objective 2.
    capacity: Maximum allowed total weight.

    Returns:
    A new neighbor solution (numpy array).
    """
    base_solution = archive[0][0].copy()
    new_solution = base_solution.copy()
    new_solution[0], new_solution[1] = new_solution[1], new_solution[0]

    return new_solution
\end{lstlisting}

\end{tcolorbox}

\begin{tcolorbox}[
    colframe=black!75!blue,
    colback=white,
    sharp corners=southwest,
    title=\textbf{Bi-CVRP heuristic design task description and template program.},
    title style={font=\bfseries\color{myblue}},
    boxrule=0.6pt,
    arc=6pt,
    left=1mm, right=1mm,
    top=0.5mm, bottom=0.5mm
]

\textbf{\textcolor{myblue}{Task Description:}} 
You are solving a Bi-objective Capacitated Vehicle Routing Problem (Bi-CVRP), where a single depot and multiple customers are located in 2D space. 
Each customer has a positive demand, and all vehicles in the fleet have identical capacity limits. 
The objective is to construct a set of routes, each starting and ending at the depot, such that:

\begin{itemize}
    \item all customers are served,
    \item vehicle capacities are not exceeded on any route,
    \item two conflicting objectives are minimized:
    \begin{itemize}
        \item total travel distance across all routes,
        \item makespan (the length of the longest individual route).
    \end{itemize}
\end{itemize}

Each solution in the archive is represented as a list of NumPy arrays, where each array denotes a single route 
(starting and ending at depot index 0), and is paired with a tuple of two objective values \texttt{(total\_distance, makespan)}. 

Your task is to implement a function named \texttt{select\_neighbor} that selects one promising solution from the archive 
and applies a novel or hybrid local search operator to generate a feasible neighbor solution. 
You must ensure that vehicle capacity constraints are respected. 

Please perform an intelligent random selection among solutions that show potential for local improvement. 
Go beyond standard approaches to develop a method that yields higher-quality solutions across both objectives. 
The function should return the new neighbor solution.

\vspace{1em}
\textbf{\textcolor{myblue}{Template Program:}}

\begin{lstlisting}[language=CustomPython]
import numpy as np
from typing import List, Tuple
import random 

def select_neighbor(
    archive: List[Tuple[np.ndarray, Tuple[float, float]]],
    coords: np.ndarray,
    demand: np.ndarray,
    distance_matrix: np.ndarray,
    capacity: float
) -> np.ndarray:
    """
    Select a promising solution from the archive and generate a neighbor solution from it.
    Args:
        archive: A list of tuples, where each tuple contains:
            - solution: A list of numpy arrays, each representing a vehicle route. 
                        Each route starts and ends at the depot (node index 0), e.g., [0, 3, 5, 0].
            - objective: A tuple of two float values (total_distance, makespan), 
                        representing the two objective values of the solution.
        
        coords: A numpy array of shape (n_nodes, 2), representing (x, y) coordinates of each node (depot + customers).
        demand: A numpy array of shape (n_nodes,), where demand[i] is the demand of node i. The depot has demand 0.
        distance_matrix: A numpy array of shape (n_nodes, n_nodes), where [i][j] is the Euclidean distance between node i and j.
        capacity: A float representing the maximum capacity of each vehicle.

    Returns:
        A new neighbor solution.
    """
    base_solution = archive[0][0].copy()
    new_solution = base_solution.copy()

    return new_solution
\end{lstlisting}

\end{tcolorbox}

\subsection{Initialization prompt}
\label{e2}
In our experiments, all initial heuristics are generated by LLMs without requiring any expert-crafted designs. The LLMs are prompted with the heuristic design task and instructed to produce new heuristics by first providing a textual description, followed by a corresponding Python implementation. This process is repeated $N$ times to obtain $N$ distinct initial heuristics.

\begin{tcolorbox}[
    colframe=purple!75!blue,
    colback=white,
    sharp corners=southwest,
    colbacktitle=purple!70!blue,
    title=\textbf{Prompt for population initialization.},
    title style={font=\bfseries\color{myblue}},
    boxrule=0.6pt,
    arc=6pt,
    left=1mm, right=1mm,
    top=0.5mm, bottom=0.5mm
]

You are an expert in the domain of optimization heuristics helping to design heuristics that can effectively solve optimization problems.\\

\textcolor{myblue}{\texttt{\{Task Description\}}}

1. First, describe your new algorithm and main steps in one long, detail sentence. The description must be inside within boxed \{\{\}.\\
2. Next, implement the following Python function:

\textcolor{myblue}{\texttt{\{Template Program\}}} \\

Check syntax, code carefully before returning the final function. Do not give additional explanations.

\end{tcolorbox}

\noindent Here, \textcolor{myblue}{\texttt{Task Description}} and \textcolor{myblue}{\texttt{Template Program}} are defined in \textbf{E.1}.

\subsection{Semantic clustering prompt}
\label{e3}
We employ the following prompt to instruct the LLM to analyze and cluster a list of heuristic code snippets based on their semantic logic. Each snippet corresponds to a heuristic generated in earlier stages and is represented as a Python function. The LLM is asked to group these snippets such that heuristics with similar behavior or design principles fall into the same cluster. This enables automated semantic analysis and categorization without manual labeling.
The expected output is a JSON object, where each key corresponds to a cluster ID (as a string), and each value is a list of indices indicating the heuristics belonging to that cluster.

To improve efficiency and avoid redundant computation, we cache the clustering results for each grid cell along with its neighboring cells. These cached results can be reused in subsequent runs, substantially reducing overall processing time.

\begin{tcolorbox}[
    colframe=teal!75!blue,
    colback=white,
    sharp corners=southwest,
    colbacktitle=teal!70!blue,
    title=\textbf{Prompt for Grouping Code Snippets},
    title style={font=\bfseries\color{mygreen}},
    boxrule=0.6pt,
    arc=6pt,
    left=1mm, right=1mm,
    top=0.5mm, bottom=0.5mm
]
You are an expert in the domain of optimization heuristics helping to design heuristics that can effectively solve optimization
problems. \\
I have \{len(codes)\} code snippets as follows: \\
\texttt{<Code>: ...} \\
... \\

Analyze the logic of all the given code snippets carefully. Then group the snippets into clusters where each group contains codes with similar logic. Return the result as a JSON object where the keys are the group indices and the values are lists of code indices that belong to each group.\\

For example:  
\begin{verbatim}
{
  "1": [0, 2, 4],
  "2": [1, 3],
  "3": [5]
}
\end{verbatim}

\end{tcolorbox}
\clearpage
\subsection{Feedback reflection prompt}
\label{e4}
We use this prompt to guide the LLM in synthesizing a new heuristic by reflecting on a set of existing candidate heuristics. Each candidate is provided in code form, and the LLM is instructed to analyze common patterns, identify shared strengths and recurring weaknesses, and ultimately propose a single hybrid or improved strategy.
\begin{tcolorbox}[
    colframe=red!75!blue,
    colback=white,
    sharp corners=southwest,
    colbacktitle=red!75!blue,
    title=\textbf{Prompt for heuristic synthesis.},
    title style={font=\bfseries\color{myblue}},
    boxrule=0.6pt,
    arc=6pt,
    left=1mm, right=1mm,
    top=0.5mm, bottom=0.5mm
]

You are an expert in the domain of optimization heuristics helping to design heuristics that can effectively solve optimization problems.\\

\textcolor{myblue}{\texttt{\{Task Description\}}}

I have \textcolor{myblue}{\texttt{\{len(indivs)\}}} existing algorithms with their codes as follows:\\
\noindent\textcolor{myred}{\texttt{<Code>}}: \quad \texttt{...} \\[0.5em]
\texttt{...}

Please carefully analyze all of the above algorithms. Your task is to synthesize their ideas, identify recurring patterns, and point out opportunities for improvement.\\

Your output should be a \textbf{Suggestions} section, where you:
\begin{itemize}
    \item Summarize key strengths shared across the implementations.
    \item Identify limitations or blind spots that appear in multiple codes.
    \item Propose hybrid or improved strategies that integrate strengths and overcome shortcomings, in a feasible running time.
\end{itemize}

\textbf{Output format:}

---

\textcolor{mygreen}{Suggestions:
Write only one proposed hybrid or improved strategy that integrates strengths and overcomes shortcomings here.}

---

Do not include any explanations, summaries, or new algorithms outside of this section.

\end{tcolorbox}
\noindent Here, \textcolor{myblue}{\texttt{len(indivs)}} is the number of selected heuristics and \textcolor{myred}{\texttt{<Code>}} are their corresponding Python implementations and descripions, respectively.
\clearpage
\subsection{Crossover prompt}
\label{e5}

\gls{mpage} adopts search operators from EoH \citep{liu2024eoh}, each implemented via LLMs.
Beyond this standard structure, we optionally include a reflective feedback segment, highlighted in red, which is derived from the synthesis output of \textbf{E.4}. This feedback provides suggestions for improvement based on analysis of previously generated heuristics. Importantly, not all operators incorporate this component: mutation operators are generated without feedback, while crossover operators may include it with a certain probability $\rho$. This probabilistic inclusion aims to balance solution quality with computational efficiency, by reducing the frequency of costly LLM inference.

\begin{tcolorbox}[
    colframe=yellow!75!blue,
    colback=white,
    sharp corners=southwest,
    colbacktitle=yellow!75!blue,
    title=\textbf{E1 Operator.},
    title style={font=\bfseries\color{myblue}},
    boxrule=0.6pt,
    arc=6pt,
    left=1mm, right=1mm,
    top=0.5mm, bottom=0.5mm
]

I have \texttt{\{len(indivs)\}} existing algorithms with their codes as follows: \\
\noindent\texttt{<Code>}: \quad \texttt{...} \\[0.5em]
\texttt{...}

\vspace{1mm}

Analyze the logic of all the given code snippets carefully. Then identify the two code snippets whose logic is most different from each other and create a new algorithm that is totally different in both logic and form from both of them.

\vspace{1mm}

\textcolor{myred}{
Here are some suggestions you can refer to: \\
--- \\
Suggestions: \\
+ \{suggestions\} + \\
---
}

\vspace{1mm}

\begin{enumerate}
    \item First, describe your new algorithm and main steps in one long, detailed sentence. The description must be inside within boxed \{\{\}.
    \item Next, implement the following Python function: \\
    \textcolor{myblue}{\texttt{\{Template Program\}}}
\end{enumerate}

Check syntax, code carefully before returning the final function. Do not give additional explanations.

\end{tcolorbox}

\begin{tcolorbox}[
    colframe=yellow!75!blue,
    colback=white,
    sharp corners=southwest,
    colbacktitle=yellow!75!blue,
    title=\textbf{E2 Operator.},
    title style={font=\bfseries\color{myblue}},
    boxrule=0.6pt,
    arc=6pt,
    left=1mm, right=1mm,
    top=0.5mm, bottom=0.5mm
]

I have \texttt{\{len(indivs)\}} existing algorithms with their codes as follows: \\
\noindent\texttt{<Code>}: \quad \texttt{...} \\[0.5em]
\texttt{...}

\vspace{1mm}

\textcolor{myred}{
Here are some suggestions you can refer to: \\
--- \\
Suggestions: \\
+ \{suggestions\} + \\
---
}

\vspace{1mm}

Please help me create a new algorithm that has a totally different form from the given ones but can be motivated from them.

\begin{enumerate}
    \item Firstly, identify the common backbone idea in the provided algorithms.
    \item Secondly, based on the backbone idea, describe your new algorithm. The description must be inside within boxed \{\{\}.
    \item Thirdly, implement the following Python function: \\
    \textcolor{myblue}{\texttt{\{Template Program\}}}
\end{enumerate}

Check syntax, code carefully before returning the final function. Do not give additional explanations.

\end{tcolorbox}

\begin{tcolorbox}[
    colframe=yellow!75!blue,
    colback=white,
    sharp corners=southwest,
    colbacktitle=yellow!75!blue,
    title=\textbf{M1 Operator.},
    title style={font=\bfseries\color{myblue}},
    boxrule=0.6pt,
    arc=6pt,
    left=1mm, right=1mm,
    top=0.5mm, bottom=0.5mm
]

I have one algorithm with its code as follows. \\[0.5em]
\noindent\texttt{<Code>}: \quad \texttt{...} \\[0.5em]
\texttt{...}

\vspace{1mm}

Please assist me in creating a new algorithm that has a different form but can be a modified version of the algorithm provided. You may focus on refining either the \textit{selection phase} or the \textit{neighborhood search phase}.

\begin{enumerate}
    \item First, describe your new algorithm and main steps in one long, detailed sentence. The description must be inside within boxed \{\{\}.
    \item Next, implement the following Python function: \\
    \textcolor{myblue}{\texttt{\{Template Program\}}}
\end{enumerate}

Check syntax, code carefully before returning the final function. Do not give additional explanations.

\end{tcolorbox}

\begin{tcolorbox}[
    colframe=yellow!75!blue,
    colback=white,
    sharp corners=southwest,
    colbacktitle=yellow!75!blue,
    title=\textbf{M2 Operator.},
    title style={font=\bfseries\color{myblue}},
    boxrule=0.6pt,
    arc=6pt,
    left=1mm, right=1mm,
    top=0.5mm, bottom=0.5mm
]

I have one algorithm with its code as follows. \\[0.5em]
\noindent\texttt{<Code>}: \quad \texttt{...} \\[0.5em]
\texttt{...}

\vspace{1mm}

Please identify the main algorithm parameters and assist me in creating a new algorithm that has a different parameter setting of the score function provided. You may focus on refining either the \textit{selection phase} or the \textit{neighborhood search phase}.

\begin{enumerate}
    \item First, describe your new algorithm and main steps in one long, detailed sentence. The description must be inside within boxed \{\{\}.
    \item Next, implement the following Python function: \\
    \textcolor{myblue}{\texttt{\{Template Program\}}}
\end{enumerate}

Check syntax, code carefully before returning the final function. Do not give additional explanations.

\end{tcolorbox}

\clearpage
\section{Clustering variations analysis}
\label{cluster}

\begin{figure}[H]
    \centering
    \includegraphics[width=1\linewidth]{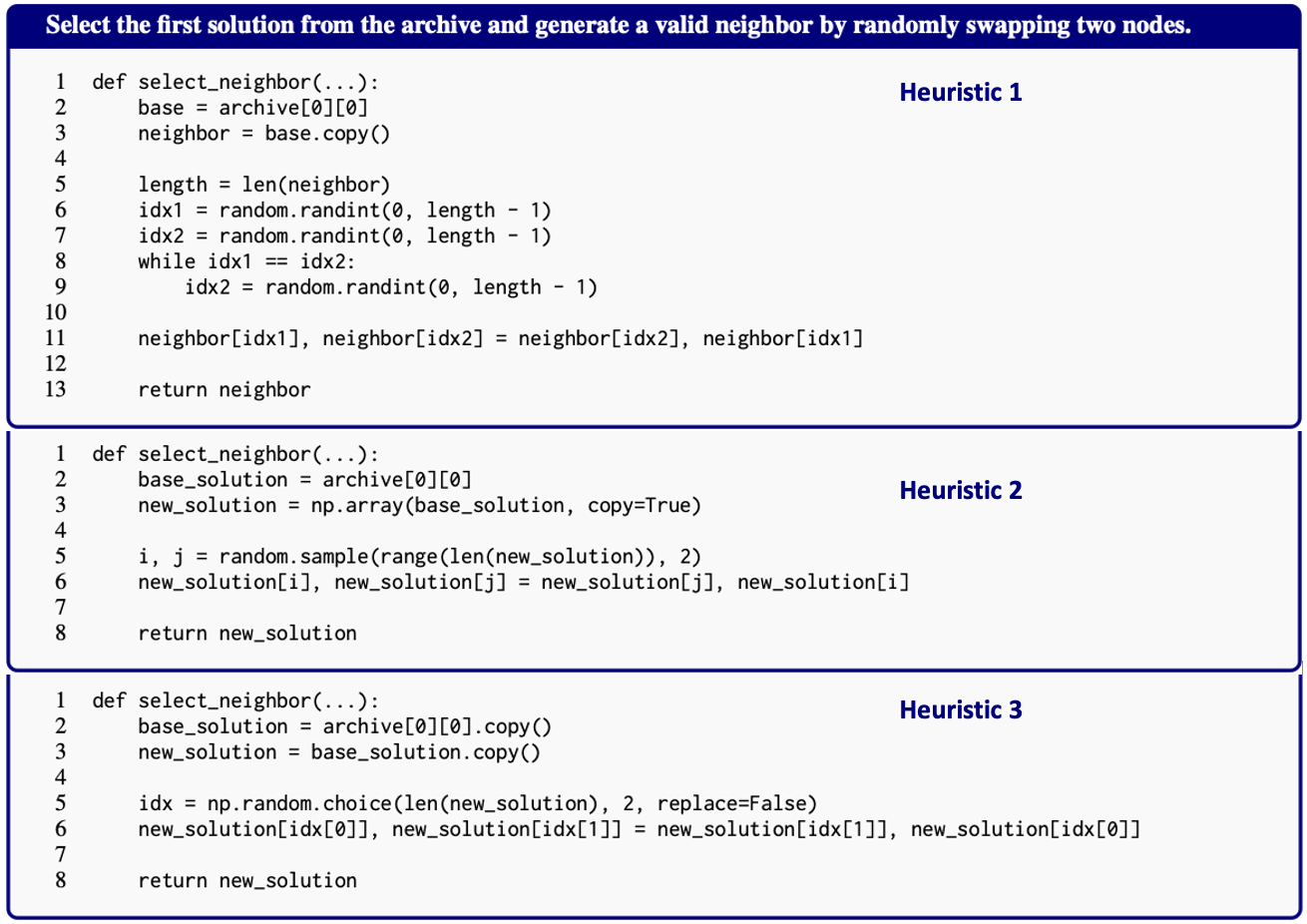}
    \caption{\small \textbf{Group Heuristics 1}: Three heuristics implementing the same logic: selecting a solution and generating a neighbor by randomly swapping two elements.}
    \label{fig:version1}
\end{figure}

\begin{figure}[H]
    \centering
    \includegraphics[width=1\linewidth]{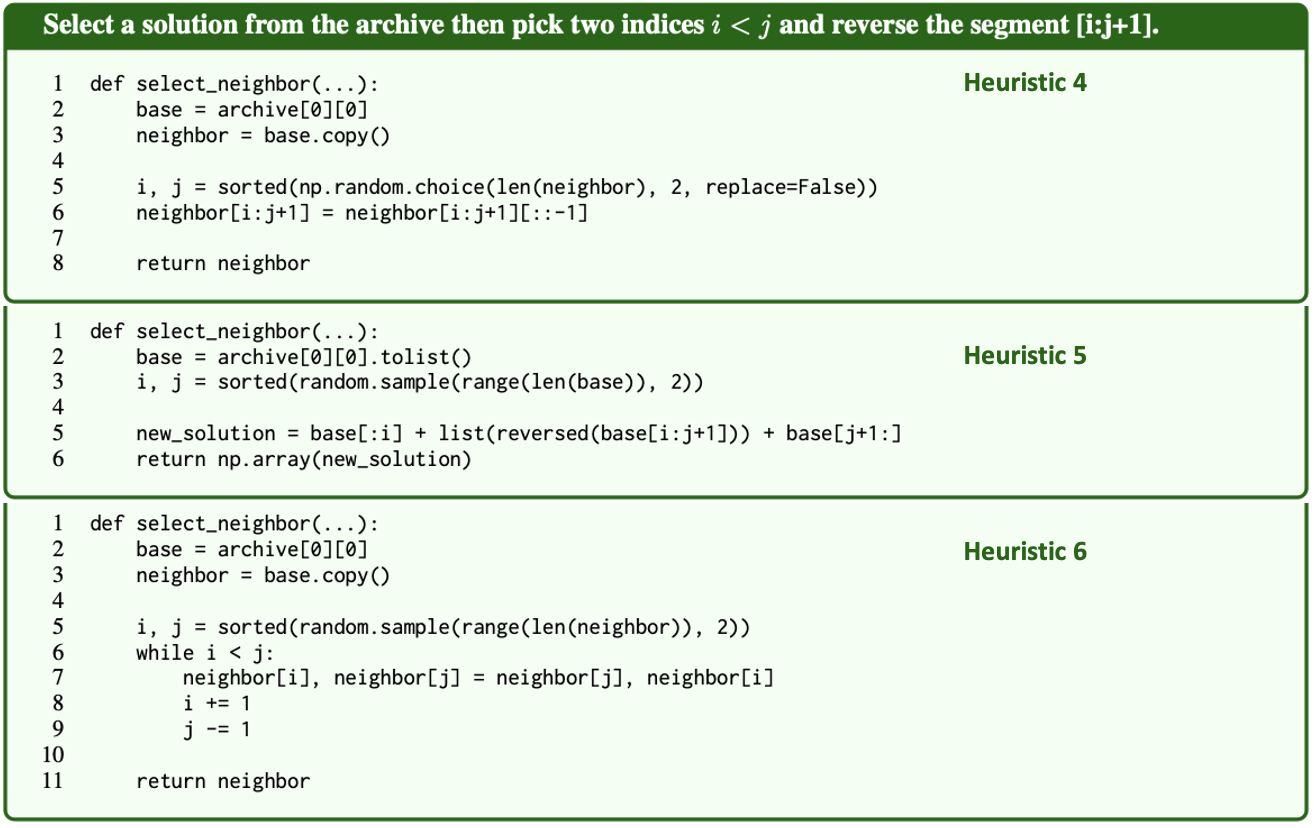}
    \caption{\small \textbf{Group Heuristics 2}: Three heuristics implementing the same logic: selecting a solution and reversing a randomly chosen segment \([i:j+1]\).}
    \label{fig:version2}
\end{figure}

\begin{figure}[H]
    \centering
    \includegraphics[width=\linewidth, height=\textheight, keepaspectratio]{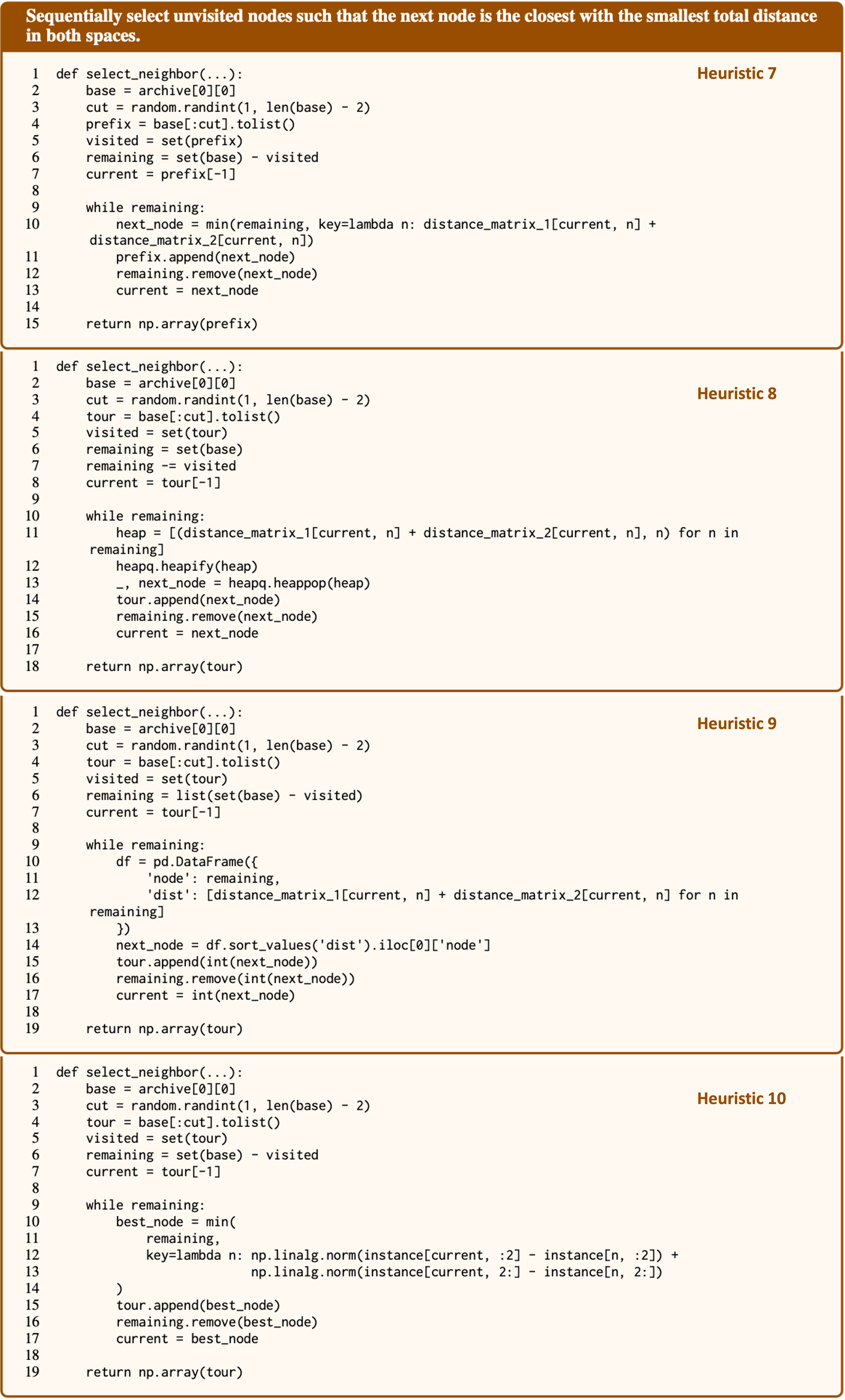}
    \caption{\small \textbf{Group Heuristics 3}: Four heuristics implementing the same logic: Sequentially select unvisited nodes such that the next node in the closet with the smallest sum of total distance.}
    \label{fig:version3}
\end{figure}

We analyze the effectiveness of different clustering methods in grouping heuristics according to their underlying logic rather than superficial code characteristics. To this end, we construct a controlled code space consisting of three ground-truth clusters, where each cluster contains multiple heuristics that implement the same functional behavior. Although all heuristics within a cluster share identical logic, they differ in implementation details such as programming constructs (e.g., use of loops vs. vectorized operations), random sampling techniques, copying mechanisms, or library functions. This design introduces syntactic and stylistic variations that challenge clustering algorithms to look beyond surface-level code differences. Illustrative examples of these heuristic groups are shown in Figures~\ref{fig:version1}--\ref{fig:version3}, where each group demonstrates the same operational intent expressed through distinct code structures.
This setup allows us to evaluate whether clustering methods can correctly recover the latent functional groupings, rather than being misled by low-level syntactic variations.


We evaluate four clustering strategies:

\begin{enumerate}
    \item \textbf{Ours (\gls{mpage})}: Leverages LLMs to semantically interpret the code and group heuristics based on inferred logical behavior.
    
    \item \textbf{SWDI-based clustering}: Converts code snippets into vector representations using a code embedding model and performs clustering based on similarity, following the setting in \citet{dat2025hsevo}.
    
    \item \textbf{K-Means}: Clusters heuristics solely based on their performance across optimization objectives, without incorporating any code-level information. The number of clusters is fixed at 3 to align with the semantic clustering setting.

    \item \textbf{AST Similarity}: Computes pairwise structural similarity between the Abstract Syntax Trees (ASTs) of the heuristics, as described in \citet{yao2025meoh}. The AST similarity score ranges from 0 to 1, where 0 indicates complete structural dissimilarity and 1 signifies syntactically identical structures.
\end{enumerate}

\begin{figure}[htbp]
    \centering

    \begin{subfigure}[b]{0.3\textwidth}
        \includegraphics[width=\textwidth]{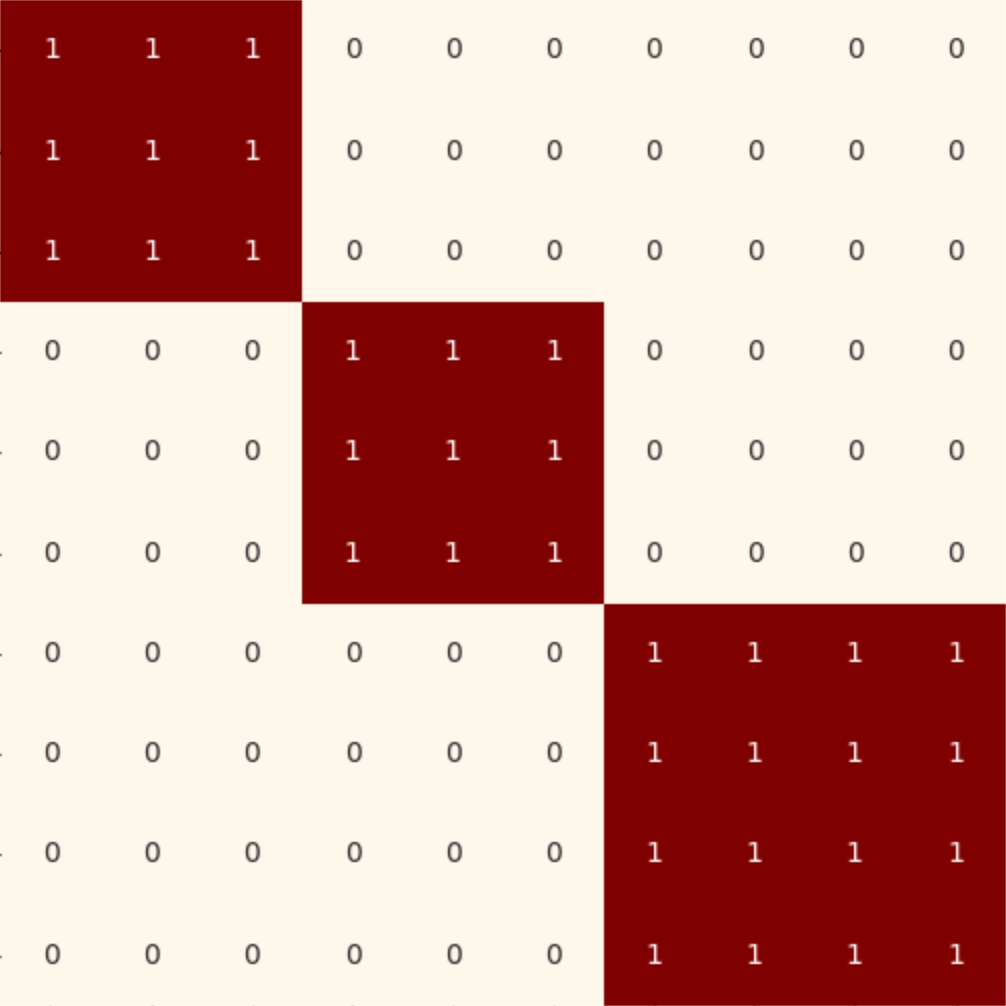}
        \caption{\small Ground Truth}
    \end{subfigure}
    \hfill
    \begin{subfigure}[b]{0.3\textwidth}
        \includegraphics[width=\textwidth]{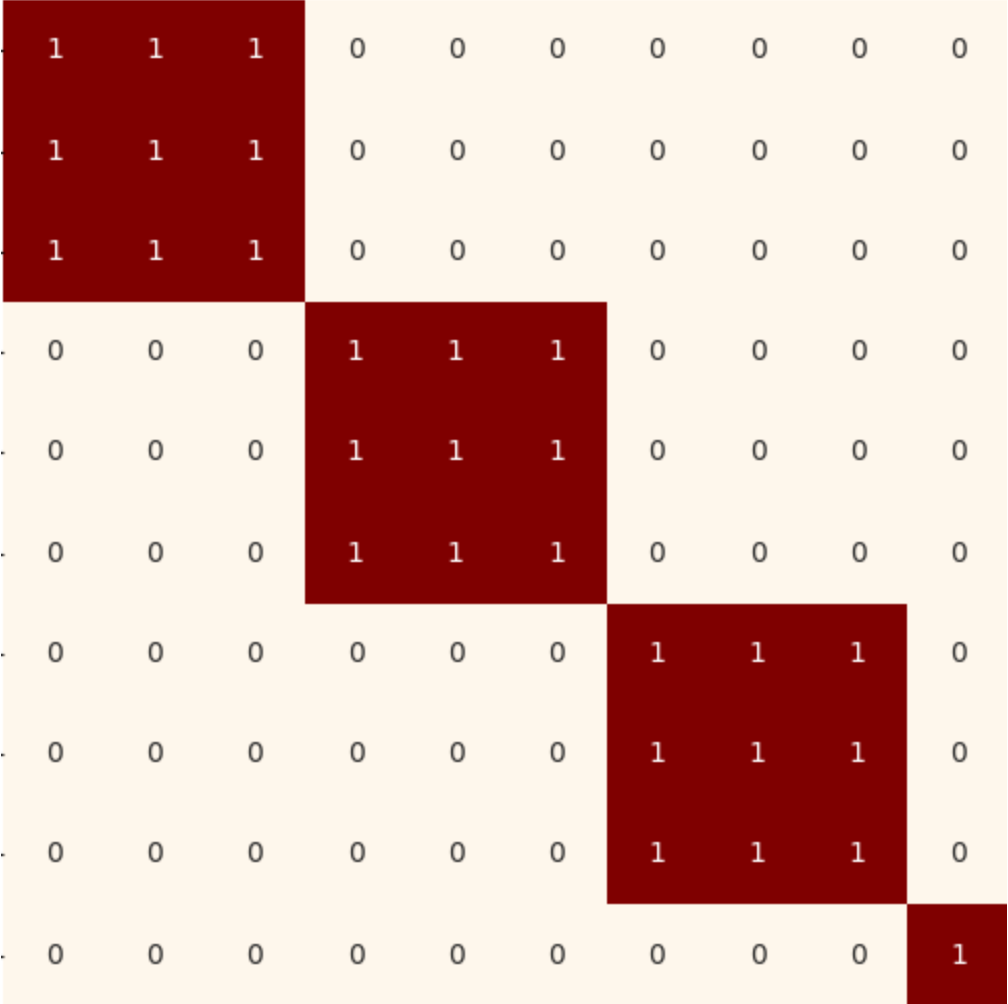}
        \caption{\small \gls{mpage}}
    \end{subfigure}
    \hfill
    \begin{subfigure}[b]{0.3\textwidth}
        \includegraphics[width=\textwidth]{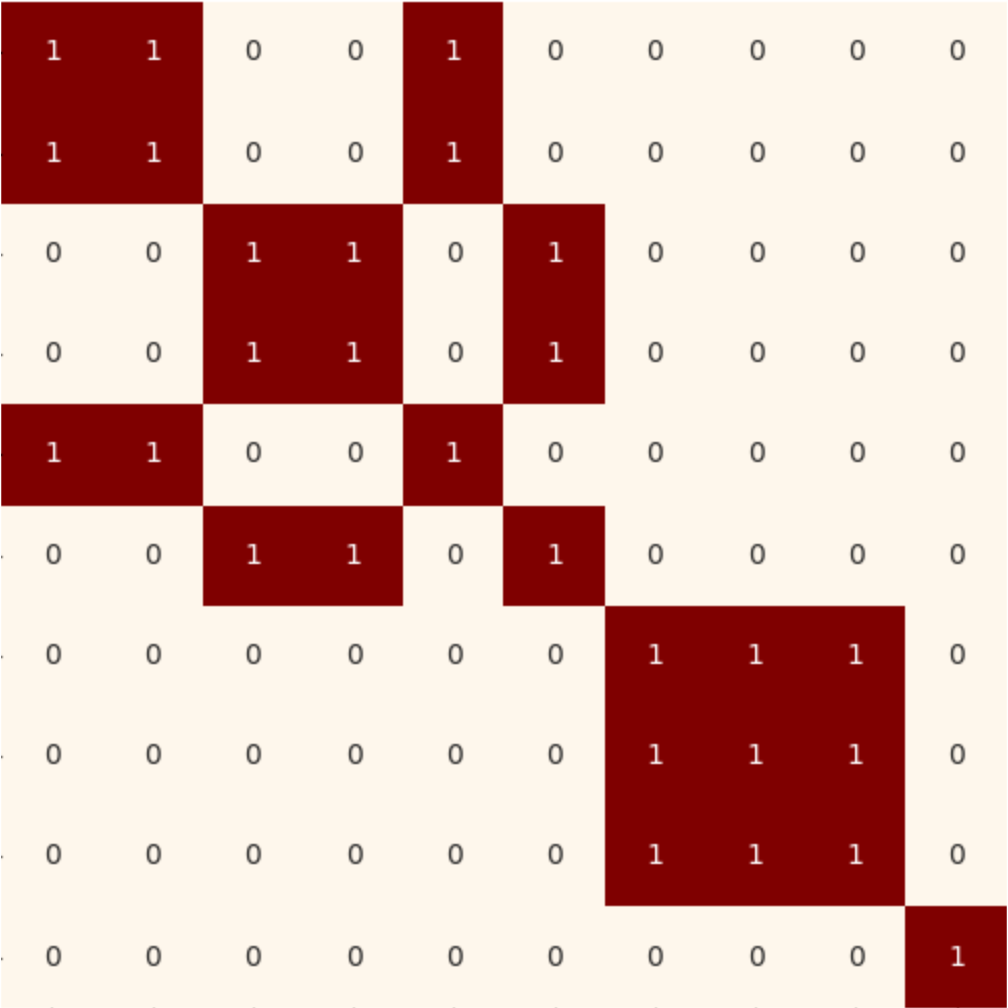}
        \caption{\small SWDI Approach}
    \end{subfigure}

    \vspace{0.5cm}

    \hspace{0.345\textwidth} 
    \begin{subfigure}[b]{0.3\textwidth}
        \includegraphics[width=\textwidth]{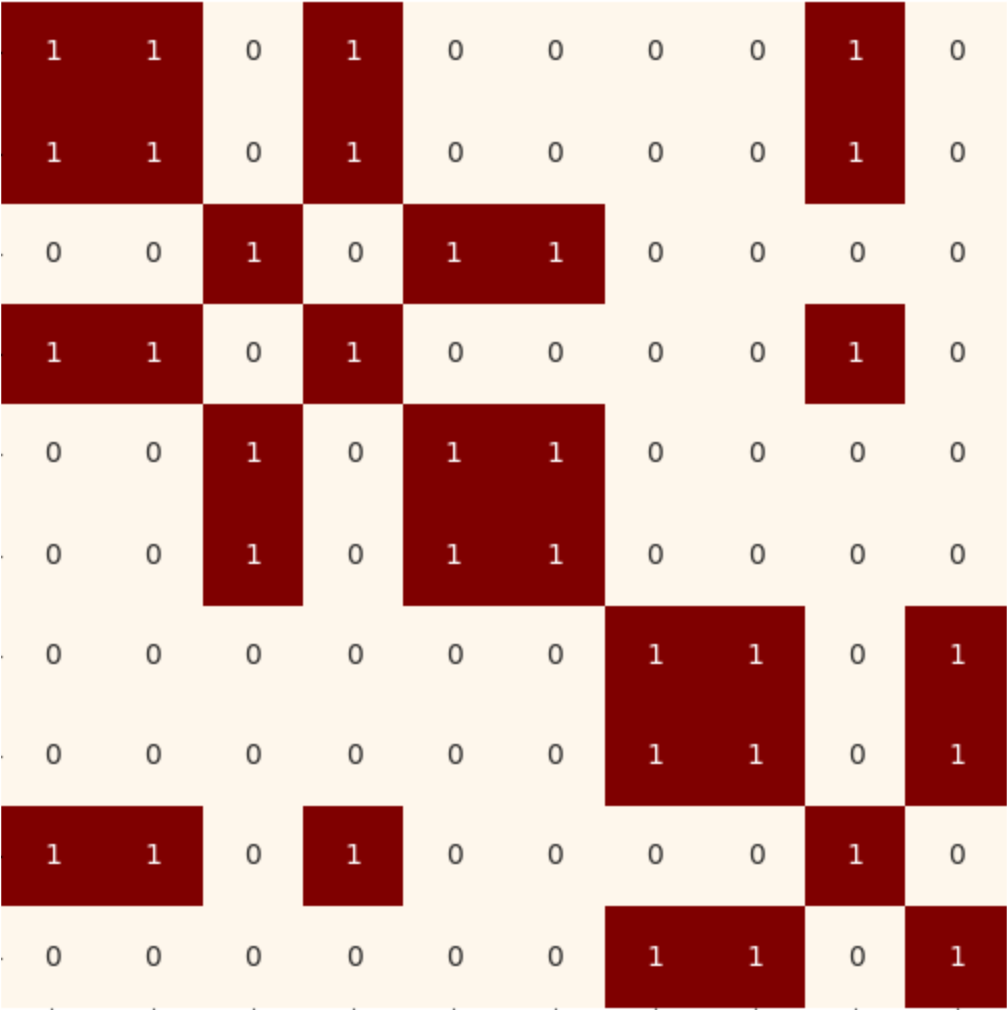}
        \caption{\small K-Means}
    \end{subfigure}
    \hfill
    \begin{subfigure}[b]{0.3\textwidth}
        \includegraphics[width=\textwidth]{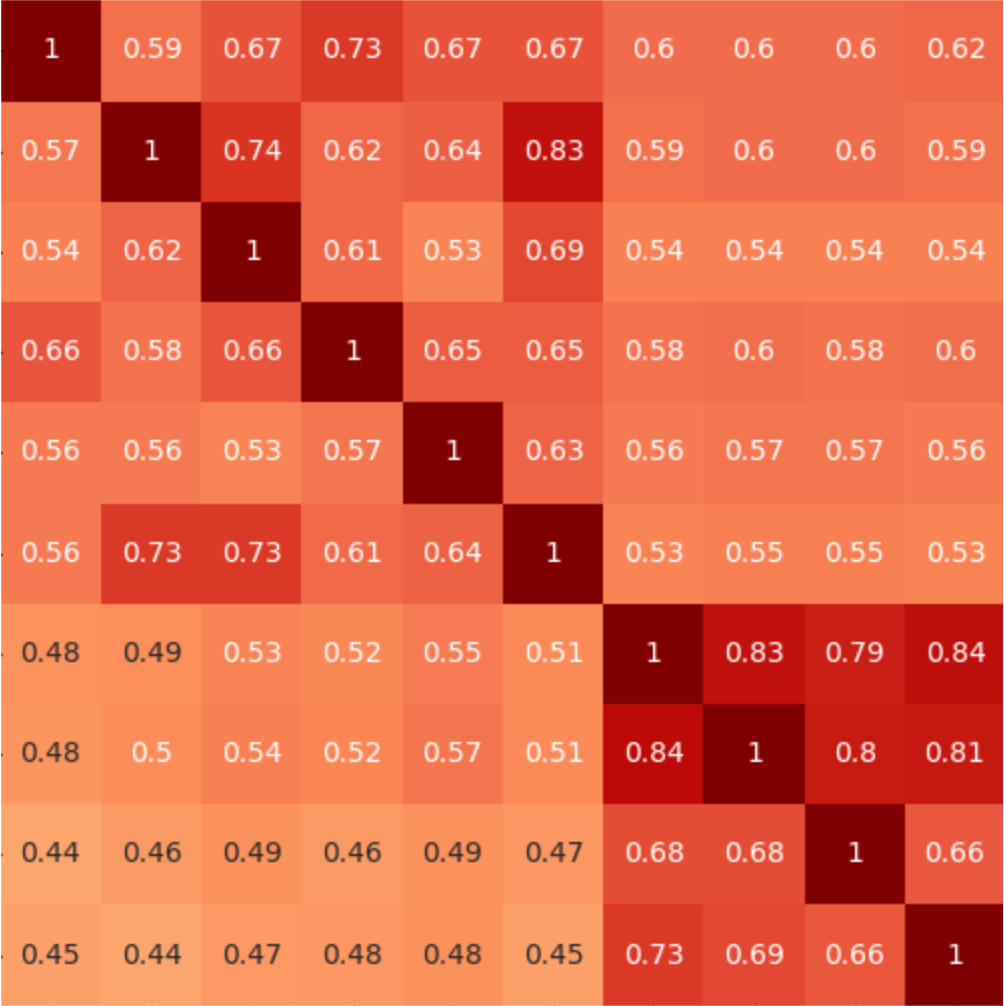}
        \caption{\small AST Similarity}
    \end{subfigure}

\caption{
\small Clustering consistency and similarity matrices across various methods. Each heatmap represents pairwise relationships between code heuristics. In subfigures (a)--(d), a value of 1 indicates that two heuristics are assigned to the same cluster, while 0 indicates they are assigned to different clusters. (a) Ground-truth grouping by functional logic. (b) Clustering result from our method (\gls{mpage}). (c) Clustering based on SWDI. (d) K-Means clustering based on objective performance. (e) Pairwise AST similarity scores between heuristics, where higher values indicate greater syntactic similarity between the corresponding Abstract Syntax Trees.
}
    \label{fig:logic-matrix-comparison}
\end{figure}

\begin{figure}[H]
    \centering
    \begin{subfigure}[b]{0.49\textwidth}
        \centering
        \includegraphics[width=\linewidth]{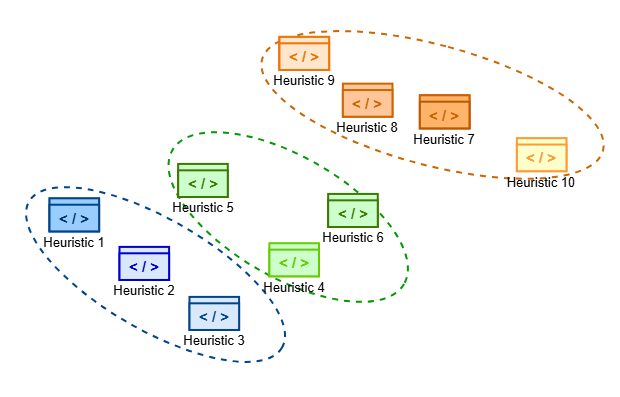}
        \caption{\small Ground Truth Clusters}
    \end{subfigure}
    \hfill
    \begin{subfigure}[b]{0.49\textwidth}
        \centering
        \includegraphics[width=\linewidth]{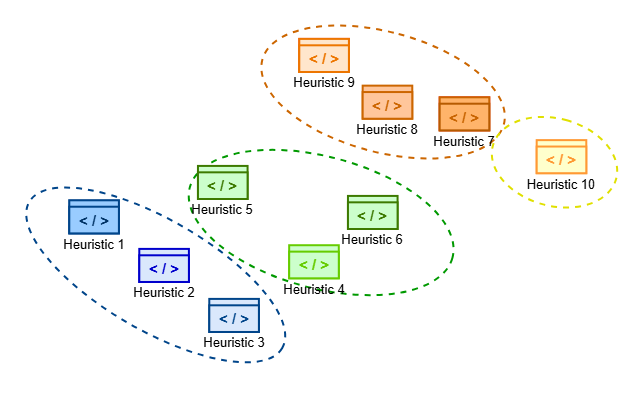}
        \caption{\small \gls{mpage} Clusters}
    \end{subfigure}


    \begin{subfigure}[b]{0.49\textwidth}
        \centering
        \includegraphics[width=\linewidth]{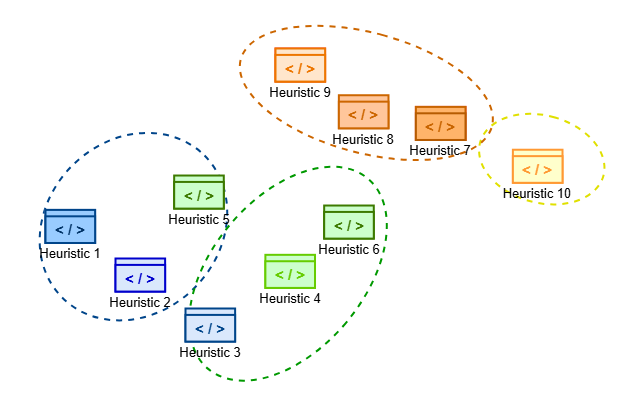}
        \caption{\small SWDI Clusters}
    \end{subfigure}
    \hfill
    \begin{subfigure}[b]{0.49\textwidth}
        \centering
        \includegraphics[width=\linewidth]{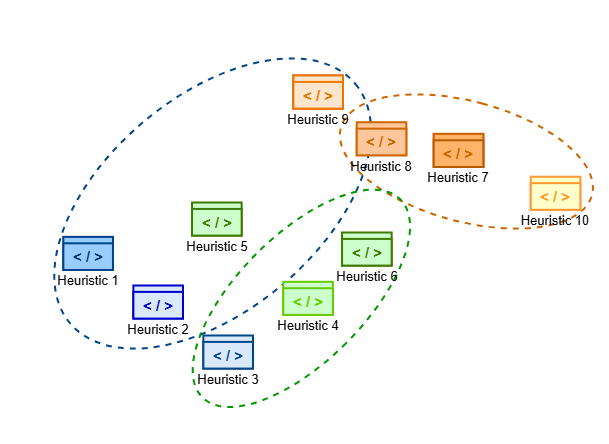}
        \caption{\small K-Means}
    \end{subfigure}

    \caption{\small Visualization of heuristic clustering results produced by different methods. \textbf{(a)} Ground truth clusters based on manual semantic annotation. \textbf{(b)} MPaGE clusters, which are generated using LLM-driven semantic interpretation of code logic. \textbf{(c)} SWDI clusters, derived from code embeddings using an embedding model. \textbf{(d)} K-Means clusters, obtained by grouping heuristics based solely on performance metrics. MPaGE most closely approximates the ground truth structure, while other methods exhibit greater semantic drift, especially in cases with subtle logic differences.}
    \label{fig:four-images}
\end{figure}

Based on Figure~\ref{fig:logic-matrix-comparison} and Figure \ref{fig:four-images}, it is evident that the \gls{mpage} method yields clustering results that most closely align with the ground-truth functional grouping. This demonstrates that LLMs, when properly utilized, are capable of understanding the semantics and intended logic of code in ways that surpass purely quantitative approaches. In contrast, the K-Means clustering method, which groups heuristics based on objective performance metrics, produces significantly less coherent clusters. This is because quantitative values, such as execution time or solution quality, do not reliably capture the underlying logic of code. Such values are often affected by stochastic behavior, system-dependent conditions, or implementation-level optimizations, leading to misleading assessments of similarity. Two heuristics implementing the same logic may yield different numerical results for reasons entirely unrelated to their semantic behavior. 

The SWDI approach, which relies on embedding representations, partially captures structural similarity (e.g., correctly grouping the last three heuristics), but overall lacks consistency. Code embeddings are heavily influenced by surface-level patterns and often fail to account for the high degree of syntactic variability introduced by LLMs. As a result, heuristics with equivalent functionality may be represented as distant points in the embedding space. The AST similarity also matrix exhibits limited discriminative power. Although ASTs reflect syntactic structure, they remain sensitive to superficial differences in code, particularly in LLM-generated heuristics, where identical logic can manifest in diverse syntactic forms. As observed in Figure 7e, the pairwise similarity scores between heuristic pairs fail to highlight or emphasize any significant structural grouping. Similarity values tend to be uniformly distributed and do not correspond well to the functional relationships among heuristics.



\begin{table}[H]
\caption{\small Performance comparison on Bi-TSP20 and Tri-TSP20 instances.}
\centering
\renewcommand{\arraystretch}{1.2}
\begin{tabular}{lcccc}
\toprule
\textbf{Method} & \multicolumn{2}{c}{\textbf{Bi-TSP20}} & \multicolumn{2}{c}{\textbf{Tri-TSP20}} \\
\cmidrule(lr){2-3} \cmidrule(lr){4-5}
 & HV $\uparrow$ & IGD $\downarrow$ & HV $\uparrow$ & IGD $\downarrow$ \\
\midrule
SWDI Cluster & 0.911 & 0.024 & 0.888 & \cellcolor{gray!50}0.098  \\
K-Means & 0.876 & 0.030 & 0.815 & 0.184 \\
AST Similarity & 0.745 & 0.184 & 0.757 & 0.203 \\
\textbf{\gls{mpage} (Ours)} & \cellcolor{gray!50}0.921 & \cellcolor{gray!50}0.013 & \cellcolor{gray!50}0.892 & 0.102 \\
\bottomrule
\end{tabular}
\label{tab:hv-igd-comparison}
\end{table}

To evaluate the effectiveness of different strategies in guiding parent selection for heuristic synthesis, we conduct experiments on two benchmark problems: Bi-TSP20 and Tri-TSP20, as shown in Table~\ref{tab:hv-igd-comparison}. For the SWDI Clustering and K-Means baselines, parent selection is performed based on cluster membership, similar to MPaGE. In contrast, the AST Similarity method samples parent pairs based on their pairwise structural similarity scores.

As shown in Table~\ref{tab:hv-igd-comparison}, MPaGE consistently outperforms all baselines across both benchmark instances. On Bi-TSP20, MPaGE achieves the highest HV (0.921) and the lowest IGD (0.013), indicating superior convergence and solution diversity. A similar pattern is observed on Tri-TSP20, where MPaGE attains the best HV (0.892) and a competitive IGD (0.102), closely approaching the best IGD achieved by SWDI Clustering (0.098). These results suggest that relying solely on AST similarity offers limited utility for parent selection, particularly when dealing with heuristics that involve complex or subtle logic structures.

\section{Designed Heuristics}
\label{design}
This section presents the best-performing heuristics generated by \gls{mpage} in terms of hypervolume and runtime across all benchmark instances.

\definecolor{mybg}{RGB}{250,250,250}
\definecolor{mykeyword}{RGB}{0,0,180}
\definecolor{mycomment}{RGB}{0,128,0}
\definecolor{mystring}{RGB}{163,21,21}
\definecolor{myidentifier}{RGB}{40,40,40}

\lstdefinelanguage{CustomPython}{
    language=Python,
    keywords={def, return, if, for, in, range, len, import, from, as, class, else, elif, True, False, None},
    keywordstyle=\color{mykeyword}\bfseries,
    identifierstyle=\color{myidentifier},
    commentstyle=\color{mycomment}\itshape,
    stringstyle=\color{mystring},
    showstringspaces=false,
    basicstyle=\ttfamily\small,
    morecomment=[l][\color{mycomment}]{\#}
}

\subsection*{Bi-TSP}
\noindent\textbf{\textcolor{blue}{\# Best hypervolume}}  \\
\hspace{1em}\textcolor{gray}{\# score = [\textcolor{red}{0.584}, \textcolor{red}{0.505}]}
\begin{lstlisting}[language=CustomPython, backgroundcolor=\color{mybg}, numbers=left, numberstyle=\tiny\color{gray}]
def select_neighbor(archive: List[Tuple[np.ndarray, Tuple[float, float]]], 
                    instance: np.ndarray, 
                    distance_matrix_1: np.ndarray, 
                    distance_matrix_2: np.ndarray) -> np.ndarray:
    # Compute selection probabilities based on inverse objective values
    total_objective_value = sum(1 / (obj[0] + 1e-9) + 1 / (obj[1] + 1e-9) 
                                for _, obj in archive)
    probabilities = [(1 / (obj[0] + 1e-9) + 1 / (obj[1] + 1e-9)) / total_objective_value 
                     for _, obj in archive]
    
    # Select a solution from the archive based on computed probabilities
    selected_index = np.random.choice(len(archive), p=probabilities)
    selected_solution = archive[selected_index][0].copy()
    
    # Generate neighbor using segment reversal
    n = len(selected_solution)
    i, j = sorted(random.sample(range(n), 2))  # Choose two distinct indices
    new_solution = np.concatenate((selected_solution[:i], 
                                   selected_solution[i:j+1][::-1], 
                                   selected_solution[j+1:]))
    
    return new_solution
\end{lstlisting}

\noindent\textbf{\textcolor{blue}{\# Best runningtime}}  \\
\hspace{1em}\textcolor{gray}{\# score = [\textcolor{red}{0.395}, \textcolor{red}{0.121}]}
\begin{lstlisting}[language=CustomPython, backgroundcolor=\color{mybg}, numbers=left, numberstyle=\tiny\color{gray}]
def select_neighbor(archive: List[Tuple[np.ndarray, Tuple[float, float]]], 
                    instance: np.ndarray, 
                    distance_matrix_1: np.ndarray, 
                    distance_matrix_2: np.ndarray) -> np.ndarray:
    # Compute selection probabilities based on inverse objective values
    total_score = sum(1 / (obj[0] + 1e-9) + 1 / (obj[1] + 1e-9) 
                      for _, obj in archive)
    probabilities = [(1 / (obj[0] + 1e-9) + 1 / (obj[1] + 1e-9)) / total_score 
                     for _, obj in archive]
    
    # Select a solution from the archive based on computed probabilities
    selected_index = np.random.choice(len(archive), p=probabilities)
    selected_solution = archive[selected_index][0].copy()
    
    # Generate neighbor using node repositioning
    n = len(selected_solution)
    neighbor_solution = selected_solution.copy()
    
    # Select two distinct nodes
    idx1, idx2 = random.sample(range(n), 2)
    
    # Heuristic: move node1 right after, node2 right before
    new_position1 = (idx1 + 1) % n
    new_position2 = (idx2 - 1) % n
    
    # Apply the swap
    neighbor_solution[new_position1], neighbor_solution[new_position2] = neighbor_solution[idx1], neighbor_solution[idx2]
    
    return neighbor_solution
\end{lstlisting}

\subsection*{Tri-TSP}
\noindent\textbf{\textcolor{blue}{\# Best hypervolume}}  \\
\hspace{1em}\textcolor{gray}{\# score = [\textcolor{red}{0.359}, \textcolor{red}{0.893}]}
\begin{lstlisting}[language=CustomPython, backgroundcolor=\color{mybg}, numbers=left, numberstyle=\tiny\color{gray}]
def select_neighbor(archive: List[Tuple[np.ndarray, Tuple[float, float, float]]], 
                    instance: np.ndarray, 
                    distance_matrix_1: np.ndarray, 
                    distance_matrix_2: np.ndarray, 
                    distance_matrix_3: np.ndarray) -> np.ndarray:
    # Select a promising solution based on multi-objective performance
    archive_weights = [1 / (1 + sum(obj)) for _, obj in archive]
    weighted_archive = random.choices(archive, weights=archive_weights, k=1)[0]
    base_solution = weighted_archive[0].copy()
    
    new_solution = base_solution.copy()
    N = len(new_solution)

    # Compute performance score and set perturbation factor
    first_objective, second_objective, third_objective = weighted_archive[1]
    avg_objective = (first_objective + second_objective + third_objective) / 3
    perturbation_factor = max(0.1, 0.5 + (0.5 - avg_objective))

    # Diversify neighborhood: Swap, Reverse, or Shift
    mutation_type = random.choices(['swap', 'reverse', 'shift'], 
                                   weights=[0.5 * perturbation_factor, 
                                            0.3 * (1 - perturbation_factor), 
                                            0.2], 
                                   k=1)[0]
    
    if mutation_type == 'swap':
        idx1, idx2 = random.sample(range(1, N - 1), 2)
        new_solution[idx1], new_solution[idx2] = new_solution[idx2], new_solution[idx1]
    elif mutation_type == 'reverse':
        start_idx = random.randint(1, N - 2)
        end_idx = random.randint(start_idx + 1, N - 1)
        new_solution[start_idx:end_idx + 1] = new_solution[start_idx:end_idx + 1][::-1]
    elif mutation_type == 'shift':
        shift_idx = random.randint(1, N - 2)
        new_solution[shift_idx], new_solution[shift_idx + 1] = new_solution[shift_idx + 1], new_solution[shift_idx]

    # Adaptive large perturbation for good solutions
    if avg_objective < 0.5:
        perturb_indices = random.sample(range(1, N - 1), k=min(3, N - 2))
        random.shuffle(perturb_indices)
        for i in range(len(perturb_indices) - 1):
            new_solution[perturb_indices[i]], new_solution[perturb_indices[i + 1]] = (
                new_solution[perturb_indices[i + 1]], new_solution[perturb_indices[i]]
            )

    # Additional large shuffle with some probability
    if random.random() < 0.3:
        additional_indices = random.sample(range(1, N - 1), k=3)
        random.shuffle(additional_indices)
        new_solution[additional_indices[0]], new_solution[additional_indices[1]], new_solution[additional_indices[2]] = (
            new_solution[additional_indices[1]], new_solution[additional_indices[2]], new_solution[additional_indices[0]]
        )

    return new_solution
\end{lstlisting}

\vspace{0.5em}

\noindent\textbf{\textcolor{blue}{\# Best running time}}  \\
\hspace{1em}\textcolor{gray}{\# score = [\textcolor{red}{0.335}, \textcolor{red}{0.476}]}
\begin{lstlisting}[language=CustomPython, backgroundcolor=\color{mybg}, numbers=left, numberstyle=\tiny\color{gray}]
def select_neighbor(archive: List[Tuple[np.ndarray, Tuple[float, float, float]]], 
                    instance: np.ndarray, 
                    distance_matrix_1: np.ndarray, 
                    distance_matrix_2: np.ndarray, 
                    distance_matrix_3: np.ndarray) -> np.ndarray:
    # Select a solution with good objective sum
    selected_solution, _ = random.choices(
        archive, 
        weights=[1 / (obj[0] + obj[1] + obj[2]) for _, obj in archive], 
        k=1
    )[0]
    
    # Apply a simple swap
    neighbor_solution = selected_solution.copy()
    n = len(neighbor_solution)
    i, j = random.sample(range(n), 2)
    neighbor_solution[i], neighbor_solution[j] = neighbor_solution[j], neighbor_solution[i]
    
    return neighbor_solution
\end{lstlisting}

\subsection*{Bi-KP}
\noindent\textbf{\textcolor{blue}{\# Best hypervolume}}  \\
\hspace{1em}\textcolor{gray}{\# score = [\textcolor{red}{0.348}, \textcolor{red}{0.130}]}
\begin{lstlisting}[language=CustomPython, backgroundcolor=\color{mybg}, numbers=left, numberstyle=\tiny\color{gray}]
def select_neighbor(archive: List[Tuple[np.ndarray, Tuple[float, float]]], 
                    weight_lst: np.ndarray, 
                    value1_lst: np.ndarray, 
                    value2_lst: np.ndarray, 
                    capacity: float) -> np.ndarray:
    # Select a solution with high objective sum
    selected_pair = random.choices(
        archive, 
        weights=[sol[1][0] + sol[1][1] for sol in archive], 
        k=1
    )[0]
    
    base_solution = selected_pair[0].copy()
    new_solution = base_solution.copy()
    current_weight = np.dot(new_solution, weight_lst)
    
    selected_indices = np.where(new_solution == 1)[0]
    unselected_indices = np.where(new_solution == 0)[0]

    # Prioritize swapping items with high profit-to-weight ratio
    if selected_indices.size > 0 and unselected_indices.size > 0:
        profit_to_weight = (value1_lst[selected_indices] + value2_lst[selected_indices]) / weight_lst[selected_indices]
        sorted_selected_indices = selected_indices[np.argsort(profit_to_weight)[::-1]]
        
        for _ in range(5):  # Try multiple swaps for improvement
            selected_idx = random.choice(sorted_selected_indices)
            unselected_idx = random.choice(unselected_indices)

            new_solution[selected_idx] = 0
            new_solution[unselected_idx] = 1
            
            if np.dot(new_solution, weight_lst) <= capacity:
                return new_solution  # Accept valid neighbor
            
            # Revert swap if constraint violated
            new_solution[selected_idx] = 1
            new_solution[unselected_idx] = 0

    # Fallback: perturb multiple items
    num_toggles = random.randint(2, 4)
    for _ in range(num_toggles):
        idx = random.randint(0, len(base_solution) - 1)
        if new_solution[idx] == 1:
            new_solution[idx] = 0
        else:
            if current_weight + weight_lst[idx] <= capacity:
                new_solution[idx] = 1
                current_weight += weight_lst[idx]

    # Final validation: enforce capacity
    while np.dot(new_solution, weight_lst) > capacity:
        deselect_idx = random.choice(np.where(new_solution == 1)[0])
        new_solution[deselect_idx] = 0

    return new_solution
\end{lstlisting}

\vspace{0.5em}

\noindent\textbf{\textcolor{blue}{\# Best running time}}  \\
\hspace{1em}\textcolor{gray}{\# score = [\textcolor{red}{0.319}, \textcolor{red}{0.074}]}
\begin{lstlisting}[language=CustomPython, backgroundcolor=\color{mybg}, numbers=left, numberstyle=\tiny\color{gray}]
def select_neighbor(archive: List[Tuple[np.ndarray, Tuple[float, float]]], 
                    weight_lst: np.ndarray, 
                    value1_lst: np.ndarray, 
                    value2_lst: np.ndarray, 
                    capacity: float) -> np.ndarray:
    # Randomly sample a base solution
    selected_solution, _ = random.choice(archive)
    neighbor_solution = selected_solution.copy()
    num_items = len(weight_lst)
    
    # Flip 1 to 3 random items with feasibility check
    for _ in range(random.randint(1, 3)):
        item_index = random.randint(0, num_items - 1)
        neighbor_solution[item_index] = 1 - neighbor_solution[item_index]
        
        # Undo flip if capacity exceeded
        while np.dot(neighbor_solution, weight_lst) > capacity:
            neighbor_solution[item_index] = 1 - neighbor_solution[item_index]
            item_index = random.randint(0, num_items - 1)
            neighbor_solution[item_index] = 1 - neighbor_solution[item_index]

    return neighbor_solution
\end{lstlisting}

\subsection*{Bi-CVRP}
\noindent\textbf{\textcolor{blue}{\# Best hypervolume}}  \\
\hspace{1em}\textcolor{gray}{\# score = [\textcolor{red}{0.417}, \textcolor{red}{0.158}]}
\begin{lstlisting}[language=CustomPython, backgroundcolor=\color{mybg}, numbers=left, numberstyle=\tiny\color{gray}]
def select_neighbor(archive: List[Tuple[np.ndarray, Tuple[float, float]]], 
                    coords: np.ndarray, 
                    demand: np.ndarray, 
                    distance_matrix: np.ndarray, 
                    capacity: float) -> np.ndarray:
    # Select a solution with best makespan (2nd objective)
    best_solution, _ = min(archive, key=lambda x: x[1][1])
    neighbor_solution = [np.copy(route) for route in best_solution]
    
    # Choose two distinct routes
    route_from_index = np.random.choice(len(neighbor_solution))
    route_to_index = np.random.choice([i for i in range(len(neighbor_solution)) if i != route_from_index])

    if len(neighbor_solution[route_from_index]) > 2 and len(neighbor_solution[route_to_index]) > 2:
        # Select customers (excluding depot)
        customer_from_index = np.random.randint(1, len(neighbor_solution[route_from_index]) - 1)
        customer_to_index = np.random.randint(1, len(neighbor_solution[route_to_index]) - 1)
        
        customer_from = neighbor_solution[route_from_index][customer_from_index]
        customer_to = neighbor_solution[route_to_index][customer_to_index]

        # Swap the customers
        neighbor_solution[route_from_index][customer_from_index] = customer_to
        neighbor_solution[route_to_index][customer_to_index] = customer_from

        # Validate route demands
        demand_from = np.sum(demand[neighbor_solution[route_from_index]])
        demand_to = np.sum(demand[neighbor_solution[route_to_index]])

        if demand_from > capacity or demand_to > capacity:
            # Revert if infeasible
            neighbor_solution[route_from_index][customer_from_index] = customer_from
            neighbor_solution[route_to_index][customer_to_index] = customer_to

    return neighbor_solution
\end{lstlisting}

\vspace{0.5em}

\noindent\textbf{\textcolor{blue}{\# Best running time}}  \\
\hspace{1em}\textcolor{gray}{\# score = [\textcolor{red}{0.146}, \textcolor{red}{0.033}]}
\begin{lstlisting}[language=CustomPython, backgroundcolor=\color{mybg}, numbers=left, numberstyle=\tiny\color{gray}]
def select_neighbor(archive: List[Tuple[np.ndarray, Tuple[float, float]]], 
                    coords: np.ndarray, 
                    demand: np.ndarray, 
                    distance_matrix: np.ndarray, 
                    capacity: float) -> np.ndarray:
    # Select route with smallest makespan
    min_makespan_solution = min(archive, key=lambda x: x[1][1])
    routes = min_makespan_solution[0]

    # Choose two distinct routes
    route1_index, route2_index = np.random.choice(len(routes), 2, replace=False)
    route1 = routes[route1_index]
    route2 = routes[route2_index]

    if len(route1) > 2 and len(route2) > 2:
        # Choose customers (excluding depot)
        customer1_index = np.random.randint(1, len(route1) - 1)
        customer2_index = np.random.randint(1, len(route2) - 1)

        customer1 = route1[customer1_index]
        customer2 = route2[customer2_index]

        # Attempt to swap
        new_route1 = route1.copy()
        new_route2 = route2.copy()
        new_route1[customer1_index], new_route2[customer2_index] = customer2, customer1

        # Check if feasible
        demand1 = np.sum(demand[new_route1[1:-1]])
        demand2 = np.sum(demand[new_route2[1:-1]])

        if demand1 <= capacity and demand2 <= capacity:
            routes[route1_index] = new_route1
            routes[route2_index] = new_route2
        else:
            # Fallback: remove and insert
            idx_remove = np.random.randint(1, len(route1) - 1)
            customer = route1[idx_remove]
            new_route1 = np.delete(route1, idx_remove)

            if np.sum(demand[new_route1[1:-1]]) + demand[customer] <= capacity:
                routes[route1_index] = new_route1
                routes[route2_index] = np.insert(route2, -1, customer)

    return routes
\end{lstlisting}
\clearpage
\section{Comparison to Neural Combinatorial Optimization}
\label{NCO}

\begin{table}[ht]
\centering
\scriptsize
\setlength{\tabcolsep}{3pt}
\caption{\small Performance comparison across all benchmarks against  Neural Combinatorial Optimization methods.}
\begin{tabular}{l|
ccc|ccc|ccc|
ccc|ccc|ccc}
\toprule
\multirow{2}{*}{Method} 
& \multicolumn{3}{c|}{Bi-TSP20} & \multicolumn{3}{c|}{Bi-TSP50} & \multicolumn{3}{c|}{Bi-TSP100} 
& \multicolumn{3}{c|}{Tri-TSP20} & \multicolumn{3}{c|}{Tri-TSP50} & \multicolumn{3}{c}{Tri-TSP100} \\
& HV $\uparrow$ & Gap & Time $\downarrow$& HV $\uparrow$ & Gap & Time $\downarrow$ & HV $\uparrow$ & Gap & Time $\downarrow$
& HV $\uparrow$ & Gap & Time $\downarrow$ & HV $\uparrow$ & Gap & Time $\downarrow$ & HV $\uparrow$ & Gap & Time $\downarrow$ \\
\midrule
PMOCO         & 0.632 & 0.31 & 3.739 & 0.637 & 0.93 & \cellcolor{gray!50}5.405 & 0.699 & 1.82 & \cellcolor{gray!50}7.952 & 0.470 & 1.67 & \cellcolor{gray!50}4.448 & 0.436 & 30.68 & \cellcolor{gray!50}5.596 & 0.490 & 4.10 & \cellcolor{gray!50}9.296 \\
NHDE-P & 0.630 & 0.63 & \cellcolor{gray!50}3.373 & 0.510 & 20.68 & 9.250 & 0.703 & 1.26 & 11.907 & 0.474 & 0.83 & 14.580 & \cellcolor{gray!50}0.629 & \textbf{\underline{0.00}} & 25.123 & 0.506 & 0.97 & 65.506 \\

NHDE-M & \cellcolor{gray!50}0.634 & \textbf{\underline{0.00}} & 93.421 & \cellcolor{gray!50}0.643 & \textbf{\underline{0.00}} & 157.843 & \cellcolor{gray!50}0.712 & \textbf{\underline{0.00}} & 328.562 & 0.476 & 0.41 & 1195.237 & 0.447 & 28.93 & 1498.893 & \cellcolor{gray!50}0.511 & \textbf{\underline{0.00}} & 3612.458 \\
\midrule
\gls{mpage} (best)   & 0.629 & 0.78 & 4.429 & 0.542 & 15.70 & 6.950 & 0.442 & 37.92 & 12.719 & \cellcolor{gray!50}0.478 & \underline{0.00} & 8.112 & 0.220 & 65.02 & 14.747 & 0.109 & 78.66 & 26.394 \\
\midrule
\midrule
\multirow{2}{*}{Method} 
& \multicolumn{3}{c|}{Bi-KP50} & \multicolumn{3}{c|}{Bi-KP100} & \multicolumn{3}{c|}{Bi-KP200} 
& \multicolumn{3}{c|}{Bi-CVRP20} & \multicolumn{3}{c|}{Bi-CVRP50} & \multicolumn{3}{c}{Bi-CVRP100} \\
& HV $\uparrow$ & Gap & Time $\downarrow$ & HV $\uparrow$& Gap & Time $\downarrow$ & HV $\uparrow$ & Gap & Time $\downarrow$
& HV $\uparrow$ & Gap & Time $\downarrow$ & HV $\uparrow$ & Gap & Time $\downarrow$ & HV $\uparrow$ & Gap & Time $\downarrow$ \\
\midrule
PMOCO & \cellcolor{gray!50}0.378 & \underline{0.00} & 5.415 
      & 0.441 & 9.25 & 8.597 
      & 0.350 & 5.14 & 11.060 
      & 0.407 & 28.34 & 5.221 
      & 0.413 & 9.03 & 9.684 
      & 0.329 & 22.03 & 14.759 \\

NHDE-P & 0.371 & 1.85 & 4.681 & 0.431 & 11.31 & 8.224 & 0.359 & 2.71 & 14.092 & 0.411 & 27.64 & 3.509 & 0.432 & 4.84 & 5.981 & 0.380 & 9.95 & 10.747 \\

NHDE-M & 0.353 & 6.61 & 283.514 & 0.453 & 6.79 & 502.367 & \cellcolor{gray!50}0.369 & \textbf{\underline{0.00}} & 897.214 & 0.421 & 25.88 & 213.678 & 0.441 & 2.86 & 299.112 & 0.396 & 6.16 & 655.309 \\
\midrule
\gls{mpage} (best)   & 0.359 & 5.02 & \cellcolor{gray!50}2.647 & \cellcolor{gray!50}0.486 & \textbf{\underline{0.00}} & \cellcolor{gray!50}1.859 & 0.197 & 46.61 & \cellcolor{gray!50}2.118 & \cellcolor{gray!50}0.568 & \textbf{\underline{0.00}} & \cellcolor{gray!50}0.072 & \cellcolor{gray!50}0.454 & \textbf{\underline{0.00}} & \cellcolor{gray!50}0.191 & \cellcolor{gray!50}0.422 & \textbf{\underline{0.00}} & \cellcolor{gray!50}0.359 \\
\bottomrule
\end{tabular}
\label{tab:nco-comparison}
\end{table}

In this section, we compare our proposed \gls{mpage} method against three representative NCO-based baselines: PMOCO, NHDE-P, and NHDE-M \citep{chen2023nhde, lin2022pareto}. As shown in Table~\ref{tab:nco-comparison}, \gls{mpage} achieves the best solution quality, measured by hypervolume (HV), in 5 out of 12 benchmarks, while consistently offering substantial runtime advantages. For instance, it is over 100$\times$ faster than NHDE-M on Tri-TSP50 (14.75s and 1498.89s) and achieves a 137$\times$ speedup on Tri-TSP100. Notably, \gls{mpage} outperforms all NCO baselines in both hypervolume and runtime across all Bi-CVRP instances, demonstrating its strength not only in efficiency but also in solution quality on challenging vehicle routing problems.
Although NCO methods can achieve strong results when carefully trained, a critical limitation is their reliance on retraining from scratch whenever the instance size changes to obtain better result, which incurs significant overhead and restricts scalability. In contrast, \gls{mpage} generalizes seamlessly across problem sizes without any retraining. These results suggest that \gls{mpage} offers a compelling trade-off between quality and efficiency, combining competitive hypervolume performance with strong adaptability and fast inference, making it more suitable for practical deployment.



\end{document}